\documentclass[a4paper,fleqn]{cas-dc}

\usepackage[authoryear,longnamesfirst]{natbib}
\usepackage{algorithm} 
\usepackage{algorithmic} 

\def\tsc#1{\csdef{#1}{\textsc{\lowercase{#1}}\xspace}}
\tsc{WGM}
\tsc{QE}

\begin{document}
\let\WriteBookmarks\relax
\def\floatpagepagefraction{1}
\def\textpagefraction{.001}

\shorttitle{TCGAN-based classification and clustering framework}

\shortauthors{F. Huang and Y. Deng}

\title [mode = title]{TCGAN: Convolutional Generative Adversarial Network for Time Series Classification and Clustering}



%

\author[1]{Fanling Huang}[style=chinese]

\cormark[1]


\ead{fanlinghuang@163.com}



\author[1]{Yangdong Deng}[style=chinese]


\ead{dengyd@mail.tsinghua.edu.cn}



\affiliation[1]{organization={School of Software, Tsinghua University},
            city={Beijing},
            country={China}}

\cortext[1]{Corresponding author}



\begin{abstract}
Recent works have demonstrated the superiority of supervised Convolutional Neural Networks (CNNs) in learning hierarchical representations from time series data for successful classification. These methods require sufficiently large labeled data for stable learning, however acquiring high-quality labeled time series data can be costly and potentially infeasible.  
Generative Adversarial Networks (GANs) have achieved great success in enhancing unsupervised and semi-supervised learning. Nonetheless, to our best knowledge, it remains unclear how effectively GANs can serve as a general-purpose solution to learn representations for time series recognition, i.e., classification and clustering.
The above considerations inspire us to introduce a Time-series Convolutional GAN (TCGAN).
TCGAN learns by playing an adversarial game between two one-dimensional CNNs (i.e., a generator and a discriminator) in the absence of label information.  Parts of the trained TCGAN are then reused to construct a representation encoder to empower linear recognition methods.
We conducted comprehensive experiments on synthetic and real-world datasets.
The results demonstrate that TCGAN is faster and more accurate than existing time-series GANs. The learned representations enable simple classification and clustering methods to achieve superior and stable performance. Furthermore, TCGAN retains high efficacy in scenarios with few-labeled and imbalanced-labeled data.
Our work provides a promising path to effectively utilize abundant unlabeled time series data.
\end{abstract}




\maketitle

\section{Introduction}
\label{sec:intro}

Time series have become increasingly valuable in various domains, such as biology, meteorology, finance, medicine, and the Internet of Things (IoT). Accurately recognizing different types of time series is an important problem, but remains challenging because the time series data tend to be noisy, dynamic, and highly problem-specific \cite{bagnall2017great,ismail2019deep,javed2020benchmark,middlehurst2021hive}.

\begin{figure}[h]
  \centering
  \includegraphics[width=1.0\linewidth]{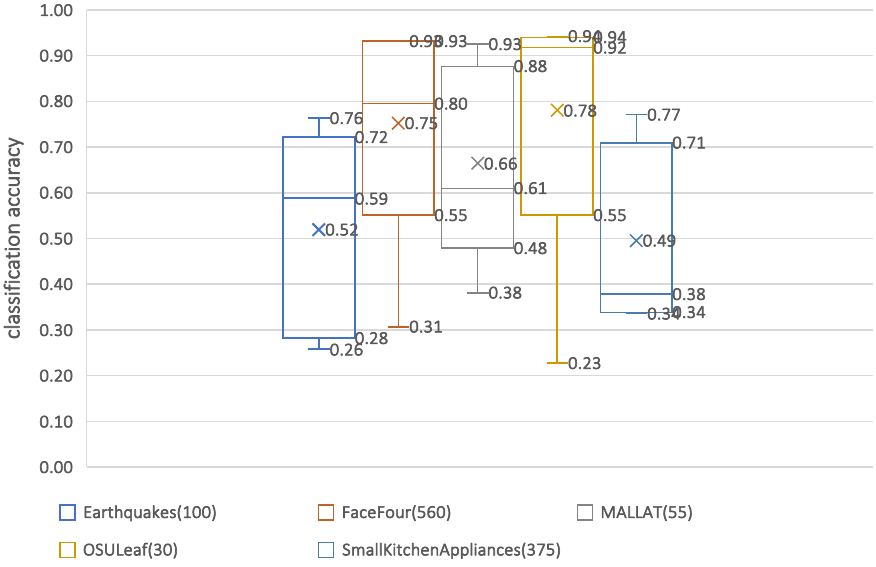}
  \caption{FCN suffers from high variance when labeled data are limited. FCN was trained on each dataset 5 times with different random seeds. The number in bracket next to the data name is the size of the training set.}
  \label{intro-dnn}
\end{figure}

In recent years, Deep Neural Networks (DNNs), in particular Convolutional Neural Networks (CNNs), have been demonstrated to be effective in time series classification due to their capability to learn hierarchical representations \cite{ismail2019deep,ismail2020inceptiontime,tang2022omni}. However, most existing methods view representation learning as an intrinsic part of a supervised DNN, which can tend to be unstable when labeled data is limited.
To illustrate this issue, we trained the Fully Convolutional Network (FCN) \cite{ismail2019deep} on 85 UCR datasets \cite{ucr15dataset} for 5 repetitions with different random seeds. We found 18 datasets with standard deviations greater than 0.1. Fig.\ref{intro-dnn} shows the five datasets with standard deviations greater than 0.2.
In practice the problem is difficult to resolve by collecting more training data because gathering high-quality labeled time series data can be costly or potentially infeasible.  
There are two main reasons for this. Firstly, human beings are less sensitive to time series, and thus the labeling process can be tedious and error-prone. Secondly, the novelties of interest (e.g., system crashes and physical lesions) in time series are innately low-probability events.

Generative models learn discriminative representations in an unsupervised manner, showing promise to alleviate the shortage of labeled data \cite{langkvist2014review}. 
In particular, Generative Adversarial Nets (GANs) have achieved great success in boosting unsupervised and semi-supervised learning \cite{goodfellow2014generative,creswell2018generative}. GANs have already found impressive applications in computer vision \cite{isola2017image,creswell2018generative,jenni2018selfsupervised}. Few studies have applied GANs for general time series generation (e.g., TimeGAN \cite{yoon2019time} and CotGAN \cite{xu2020cot}) and specific time series mining problems (e.g., anomaly detection \cite{li2018anomaly} and imputation \cite{luo2019e2gan}). However, these advanced GANs depend primarily on autoregressive models like Recurrent Neural Networks (RNNs), which learn in a sequential manner and are theoretically less efficient than CNNs that can be easily parallelized over time \cite{bai2018empirical}. 
Applying convolutional GANs for efficient sequence modeling has been explored in some task-specific studies, such as irregular sampling time series imputation \cite{ramponi2018t}, speech generation \cite{ye2020tdcgan}, human motion prediction \cite{cui2021efficient} and time series sampling \cite{dahl2021time}.  
To the best of our knowledge, it remains unclear how effectively GANs can serve as a general-purpose solution to learn representations from diverse unlabeled time series data and then support time series mining tasks like classification and clustering.
\begin{figure}[h]
  \centering
  \includegraphics[width=1.0\linewidth]{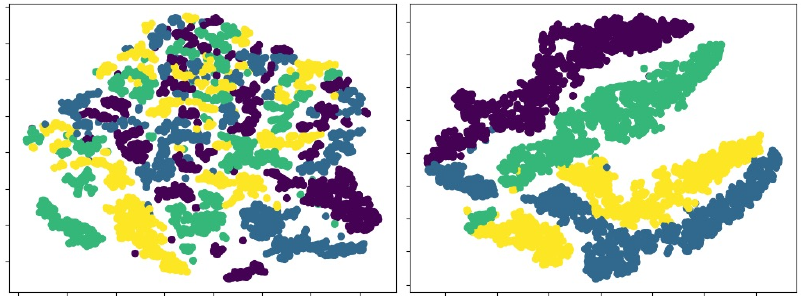}
  \caption{t-SNE plots for (left) the raw time series and (right) the TCGAN representations of the Two\_Patterns dataset. Each point corresponds to a time series. Colors annotate the four classes.}
  \label{intro-case}
\end{figure}

Fusing the two sides above, we introduce a Convolutional GAN to learn representations from unlabeled time series data, dubbed as Time-series Convolutional GANs (TCGANs). These learned representations can then be used by simple recognition models to achieve stable and superior performance.
Specifically, TCGAN consists of two one-dimensional CNNs (i.e., a generator and a discriminator). Each CNN learns through playing an adversarial game with the other. The generator attempts to generate statistically realistic time series, while the discriminator endeavors to distinguish real time series from the generated fake ones.
Parts of the trained discriminator are then reused to form an independent representation encoder. As illustrated in Fig.~\ref{intro-case}, the encoder is capable of transforming raw time series into an alternative data space where instances of different classes can be more easily differentiated.

The major contributions of this paper can be summarized as follows: 
\begin{itemize}
\item We propose TCGAN, a GAN-based unsupervised time series representation learning framework, which can be seamlessly used with time series classification and clustering. 
\item We conduct comprehensive experiments on synthetic datasets and 85 time series datasets from the UCR archive. The experimental results show that: 
(1) TCGAN significantly outperforms leading time-series GANs in terms of generation performance, classification accuracy, and runtime. 
(2) TCGAN representations can enable simple classifiers to obtain better accuracy than unsupervised and supervised CNNs. TCGAN classifiers can also obtain a decent level of performance even where labeled data are highly limited and imbalanced. 
(3) TCGAN representation preserves the pairwise similarities between time series, which allows the distance-based clustering method to achieve higher accuracy compared with most leading competitors
\item To motivate more attention on applying generative models, particularly GANs, to serve the time series mining community, we open the source code at \url{https://bitbucket.org/Lynn1/tcgan}. 
\end{itemize}

\section{Related work}
\label{sec:rel}

\subsection{Time series classification}
\label{sec:rel-tscla}
Time series classification is one of the most pervasive research problems in the data mining domain. Existing approaches can be divided into three categories: distance-based, feature-based, and ensemble approaches.
The distance-based method integrates a distance or similarity measure into the nearest neighbor classifier. The $1$-nearest neighbor (1NN) with Dynamic Time Warping (DTW) elastic distance established a hard-to-beat baseline \cite{lines2015time}. 
The feature-based method focuses on distilling salient representations from raw data for accurate classification. 
Conventional works have mined various time series features, for example, measures of time series characteristics (skewness, trend, seasonality, etc.) \cite{christ2018time} and salient subsequences like shapelets \cite{hills2014classification}. 
Recent DNN classifiers resort to their capability to automatically learn hierarchical representations \cite{ismail2019deep}.
There is a growing consensus that no single setup can produce universally superior performance on diverse time series datasets, prompting ensemble methods, like InceptionTime \cite{ismail2020inceptiontime}, Hive-COTE \cite{middlehurst2021hive}, Omni-Scale CNNs \cite{tang2022omni} and MultiRocket \cite{tan2022multirocket}, to pursue the State-Of-The-Art (SOTA) by exhausting and fusing diverse time series representations.

Convolutional Neural Networks (CNNs) have been demonstrated to be effective and efficient in modeling sequential data \cite{bai2018empirical,ismail2019deep,tang2022omni}. 
However, single-supervised DNN tends to be unstable in the absence of labeled data.
An ensemble strategy is often used to obtain a stable performance. For example, InceptionTime \cite{ismail2020inceptiontime} unifies five Inception networks (i.e., CNNs) to reduce the variance of evaluation results. To achieve the SOTA accuracy, \cite{tang2022omni} fuses five OS-CNNs and equips each OS-CNN with the proposed Omni-Scale block (OS-block) that integrates multiple receptive fields with different kernel sizes. 
Furthermore, some works have investigated data augmentation \cite{wang2021hierarchical}, transfer learning \cite{kashiparekh2019convtimenet}, meta learning \cite{narwariya2020meta}, and self-supervised learning \cite{eldele2022self} to make DNNs applicable in scenarios where labeled time series data is lacking.

Differently, we focus on generative models, which can learn discriminative representations in an unsupervised manner \cite{langkvist2014review} but have received less attention in the time series classification community. Specifically, we leverage GANs to learn time series representations from abundant unlabeled data. The resulting representations can enable simple classifiers to obtain more superior and stable accuracy compared with single-supervised CNNs, even where labeled data are highly limited and imbalanced.

\subsection{Time series clustering}
\label{sec:rel-tsclus}
Clustering aims to group unlabeled data by uncovering complex distributions and structures inherent in the data. 
Existing time series clustering methods could be categorized into raw-data-based and feature-based methods. 
The raw-data-based approach focuses on devising an appropriate distance or similarity measure that captures the shape features in time series. Typically, the $k$-shape algorithm proposed in \cite{paparrizos2017fast} is a superior raw-data-based algorithm that uses a normalized version of the cross-correlation as distance measure.
The feature-based approach first represents a raw time series with a feature vector and then applies a clustering algorithm on the extracted feature vectors. For example, \cite{lei2019similarity} proposes the Similarity PreservIng RepresentAtion Learning (SPIRAL) algorithm and feeds the resulting representations into a k-means algorithm with DTW or Move-Split-Merge (MSM) as a distance measure. Interested readers could refer to these time series clustering surveys \cite{aghabozorgi2015time,javed2020benchmark} for more details.
Our method is feature-based. Our method derives a transformation space where the patterns of different groups of data become apparent, and the pairwise similarities between time series are well preserved. As a result, simple clustering methods like k-means with Euclidean distance can achieve superior performance.  

\subsection{Generative Adversarial Networks (GANs)}
\label{sec:rel-gan}
Generative Adversarial Networks (GANs), firstly proposed by \cite{goodfellow2014generative}, are SOTA generative models amenable for unsupervised and semi-supervised learning. 
In the image domain, there has been a large body of research on developing new architectures \cite{radford2015unsupervised}, optimization techniques \cite{gulrajani2017improved}, and applications \cite{isola2017image} under the framework of GANs. We refer the reader to recent surveys \cite{brophy2021generative,creswell2018generative} for more information.

The development of GANs for use on time series data is still in an early stage \cite{brophy2021generative}. 
Existing works mainly focus on generating realistic time series. 
TimeGAN \cite{yoon2019time} proposes to generate and discriminate within a jointly optimized embedding space, as well as combine unsupervised adversarial training with a supervised teacher-forcing component to capture the autoregressive natures of time series.
Causal Optimal Transport GAN (COT-GAN) \cite{xu2020cot} introduces a new adversarial objective based on Causal Optimal Transport (COT) theory to model both complex spatial structures and temporal dependences in time series.
Time-series Generation by Contrastive Imitation (TimeGCI) \cite{jarrett2021time} captures both the conditional dynamics of (stepwise) transitions and the joint distribution of (multi-step) trajectories in time series.
\cite{pei2021towards} pays special attention to generating long sequences with variable lengths and missing values. 
There are also some domain-specific or task-specific attempts, for example, \cite{wiese2019quant} proposes Quant GAN for financial time series generation, \cite{li2018anomaly} works on time series anomaly detection, and \cite{luo2019e2gan} targets multivariate time series imputation.

The above time-series GANs mainly depend on autoregressive models like RNNs, which are less efficient than parallelizable CNNs \cite{bai2018empirical}. 
Some studies have applied convolutional layers for efficient sequence modeling, but they only focused on limited applications and didn't open the source code for further exploration.  
\cite{ramponi2018t} uses convolutional layers in a conditional GAN for irregular sampling time series imputation and conducts experiments on 1 synthetic and 3 UCR datasets. 
\cite{ye2020tdcgan} proposes a temporal dilated convolutional GAN for speech generation and evaluates the model on one simulated speech dataset.
\cite{cui2021efficient} focuses on human motion prediction and uses temporal convolutions in the GAN structure for efficient sequence-to-sequence modeling. 
\cite{dahl2021time} uses a GAN embedded with temporal convolutions in a bootstrap-like method for time series, and conducts experiments on a simulated AR(1) time series process and a U.S. equity dataset.
In general, it remains unclear how effective Convolutional GANs are at modeling diverse time series, particularly for classification and clustering tasks

\section{Proposed framework}
\label{sec:approach}
\begin{figure*}[h]
\centering
\includegraphics[width=0.8\linewidth]{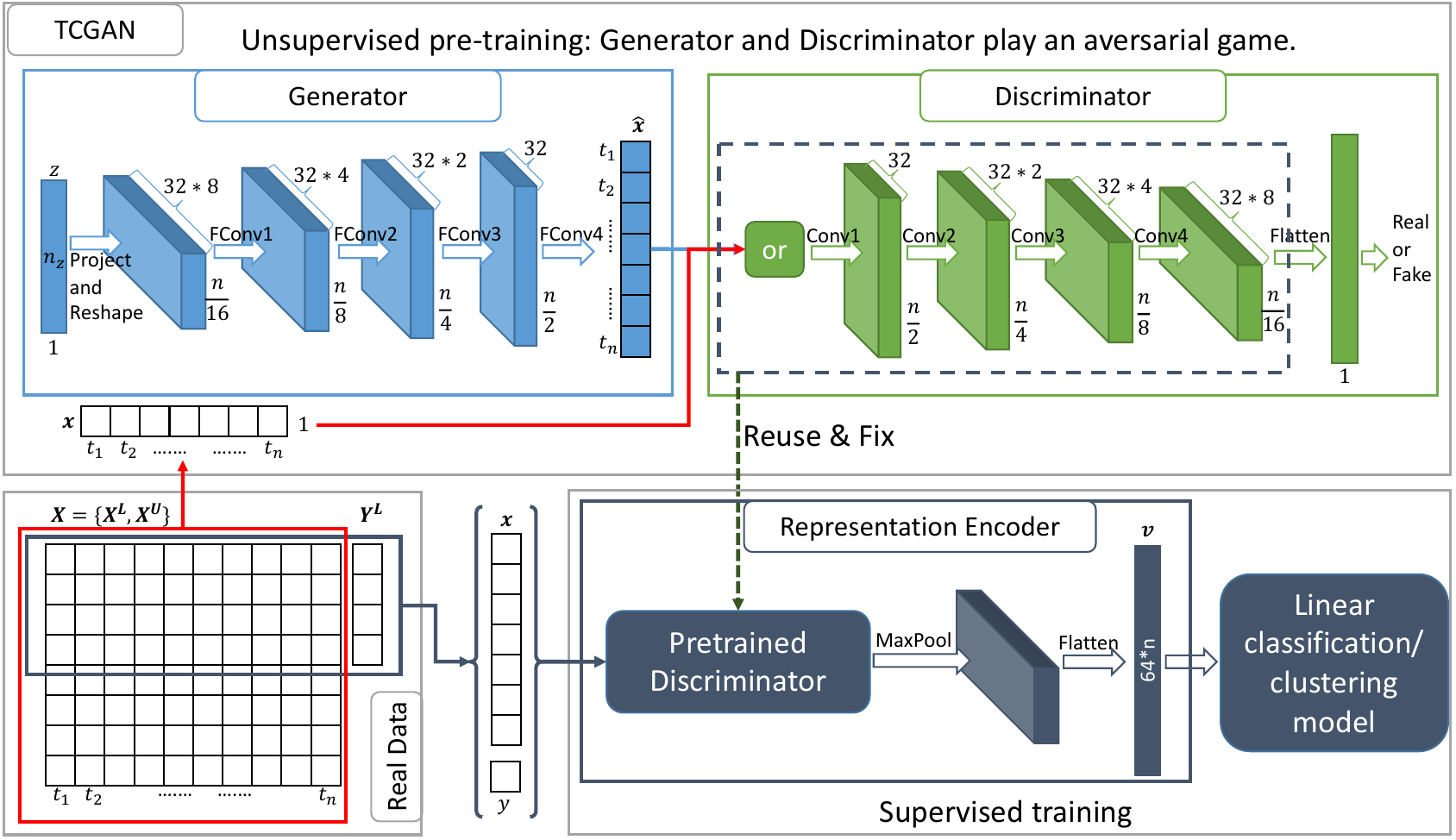}
\caption{TCGAN-based classification and clustering framework.}
\label{fig:framework}
\end{figure*}

As illustrated in Fig.~\ref{fig:framework}, we propose a Time-series Convolutional Generative Adversarial Network (TCGAN) based classification and clustering framework.
Given a training dataset containing $C$ classes. The dataset consists of a labeled subset $\mathbf{X}^{L} = \{\mathbf{x}^{i}\}_{i=1}^{L}$ with labels $\mathbf{Y}^{L} = \{y^{i}\}_{i=1}^{L}$ ($y^{i} \in \{1,2,...,C\}$) and an unlabeled subset $\mathbf{X}^{U} = \{\mathbf{x}^{i}\}_{i=1}^{U}$. Without loss of generality, we assume the first $L$ samples within $\mathbf{X}=\{\mathbf{X}^{L}, \mathbf{X}^{U}\}$ are labeled by $\mathbf{Y}^{L}$.
Each sample $\mathbf{x}^{i}$ $\in \mathbb{R}^{n*d}$ is a time series of length $n$ and number of variables $d$. For readability, the superscript $i$ to denote a sample is omitted without ambiguity. As a result, $\mathbf{x}^{i} = \mathbf{x} =[\mathbf{x}_{1}, \mathbf{x}_{2}, ..., \mathbf{x}_{t}, ..., \mathbf{x}_{n}]$, $\mathbf{x}_{t} \in \mathbb{R}^{d}$ is a vector of observations at timestamp $t$. 
First, TCGAN learns from entire training dataset $\mathbf{X}=\{\mathbf{X}^{L}, \mathbf{X}^{U}\}$ in an unsupervised manner. 
Specifically, TCGAN takes the form of a standard GAN consisting of a generator and a discriminator.
The generator, $G(\mathbf{z};\mathbf{\theta_{g}}): \mathbf{z} \in \mathbb{R}^{n_{z}} \to \mathbf{\hat{x}} \in \mathbb{R}^{n*d}$, a CNN-based sampler with parameters $\mathbf{\theta_{g}}$, learns to approximate the distribution of real data $P_{data}$. $G$ takes a random variable $\mathbf{z} $ as input that obeys a predefined prior $P_{z}$ and generates a fake time series $\mathbf{\hat{x}}$. 
The discriminator, $D(\mathbf{\widetilde{x}};\mathbf{\theta_{d}}): \mathbf{\widetilde{x}} \in \mathbb{R}^{n*d} \to p_{real} \in [0, 1]$, a binary classification CNN with parameters $\mathbf{\theta_{d}}$, outputs a single scalar $p_{real}$ representing the probability that $\mathbf{\widetilde{x}}$ comes from $P_{data}$ rather than $G(\mathbf{z};\mathbf{\theta_{g}})$. 
The generator ($G$) and discriminator ($D$) of TCGAN learn from each other by competing in an adversarial game in absence of any labeled information. 
Then, an encoder is constructed by reusing parts of the pretrained discriminator. The resulting TCGAN encoder, $E(\mathbf{x};\mathbf{\theta_{e}}): \mathbf{x} \in \mathbb{R}^{n*d} \to \mathbf{v} \in \mathbb{R}^{n_{v}}$, transforms each time series $\mathbf{x}$ into a feature vector $\mathbf{v}$ of length $n_{v}$ for an off-the-shelf linear classification or clustering model. 
In the following subsections, we will present how to build the proposed framework compatible with time series data.

\subsection{TCGAN}
We adapt our generator and discriminator architectures from an image synthesis GAN, which was developed through extensive model exploration \cite{radford2015unsupervised}. Our networks are based on the one-dimensional CNN for modeling time series data. The filters exhibit only one dimension (time) instead of two dimensions (width and height) in images. In comparison with the recent time-series GANs \cite{yoon2019time,xu2020cot} that use an autoregressive model (e.g., RNN) to capture the conditional dynamics of stepwise transitions, $p(\mathbf{x}_{t} | \mathbf{x}_{1}, ..., \mathbf{x}_{t-1}), t \in \{1,2,...,n\}$, we found modeling entire time series, $p(\mathbf{x}_{1}, ..., \mathbf{x}_{n})$, with stacking convolution layers is more efficient and effective. In general, TCGAN has the following advantages: (1) The model can see contextual information from both directions, $\mathbf{x}_{t}$ is attended to the information from $\mathbf{x}_{[:t-1]}$ and $\mathbf{x}_{[t+1:]}$. (2) Stacking convolution layers can model long dependencies and hierarchical features in time series. \cite{ismail2020inceptiontime} (3) Without modeling time series step by step, CNNs are easy to parallel \cite{bai2018empirical}. 

\subsubsection{Generator}

The generator, $G(\mathbf{z};\mathbf{\theta_{g}}): \mathbf{z} \in \mathbb{R}^{n_{z}} \to \mathbf{\hat{x}} \in \mathbb{R}^{n*d}$, maps a noise sequence $\mathbf{z} \sim P_{z}$ to a time series $\mathbf{\hat{x}}$ mimicking the real $\mathbf{x} \sim P_{data}$. $G$ has five layers and processes one batch of samples at a time. 
First, a dense layer with a ReLu activation function projects and reshapes the input to align with the following layers. Next, four one-dimensional Fractionally-strided Convolutional (FConv) layers \cite{springenberg2014striving} are applied sequentially to generate a batch of time series as output.
All FConv layers except the last one have a FConv-BatchNorm-Relu structure (the last layer just performs a FConv operation). The FConv operation uses a strided convolution to replace the deterministic spatial pooling function (e.g., maxpooling) and allows the generator to learn its own spatial up-sampling. The BatchNorm operation \cite{ioffe2015batch} is deployed to stabilize the learning process and prevent the problem of "mode collapse" in GANs. We do not use any activation function for the last layer because it tends to distort time series. In fact, we do not perform any pre- and post-processing on real or generated data except for applying $z$-normalization to the primitive time series.    
All convolutional filters share a common setting, i.e., each filter has a kernel size of $10$ and a stride of $2$. We found a large kernel size is generally helpful for time series modeling, yet reduces efficiency to some degree. Thus, we choose a moderate value of $10$. The stride of $2$ results in a 2-times up-sampling on the input after each FConv operation. 

\subsubsection{Discriminator}
The discriminator, $D(\mathbf{\widetilde{x}};\mathbf{\theta_{d}}): \mathbf{\widetilde{x}} \in \mathbb{R}^{n*d} \to p_{real} \in [0, 1]$, is a FCN for binary classification. Specifically, $D$ consists of six layers to classify each inputted time series as real or fake. 
The input is a batch of real or fake ($\mathbf{x}$/$\mathbf{\hat{x}}$) time series sequentially flowing through four one-dimensional Convolutional (Conv) layers. All Conv layers except the first one have a Convolution-BatchNorm-LeakyReLU structure (no BatchNorm operation in the first layer). All convolutional filters have the same setting, i.e., each filter has a kernel size of $10$, and a stride of $2$. The stride of $2$ results in a 2-times down-sampling of the input after each Conv operation. The output of the above process is flattened into a batch of one-dimensional feature vectors for a sigmoid layer to conduct binary classification.

\subsubsection{Training} 

Following \cite{goodfellow2014generative}, $G$ and $D$ are trained by playing a two-player minimax game with a value function defined as Eq.~\eqref{eq:min-max-game}. $D$ attempts to determine whether a sample is from $G$ or the real dataset, while $G$ tries to produce fake data (but mimicking the distribution of the real dataset) to fool $D$.
\begin{equation}
\begin{aligned}
\label{eq:min-max-game}
\underset{G}{min}\ \underset{D}{max}\ V(G, D) =  \mathbb{E}_{\mathbf{x} \sim P_{data}}[log(D(\mathbf{x}))] + \\ 
 \mathbb{E}_{\mathbf{z} \sim P_{z}}[1-log(D(G(\mathbf{z})))]
\end{aligned}
\end{equation}

Please refer to Alg.~\ref{alg:tsgan} for the algorithmic details. TCGAN is trained with unlabeled data by using the Adam algorithm \cite{kingma2014adam} with empirically chosen hyper-parameters. We follow the non-saturating training criterion for $G$, i.e., training $G$ to maximize $log(D(G(\mathbf{z})))$ instead of minimizing $log[1-D(G(\mathbf{z}))]$ so as to derive stronger gradients for $G$ in early learning \cite{goodfellow2014generative}.
A straightforward training procedure for GAN is to update $G$ and $D$ once per batch \cite{goodfellow2014generative,xu2020cot}. 
In our practical experience with time series data, we observed that $D$ usually learns much faster than $G$. A possible reason is that generating time series is more difficult than distinguishing real time series from synthetic time series. Therefore, it is hard for $G$ to extract signals for improvement from $D$ because all generated samples would be rejected by $D$ with high confidence.
In addition, we found that fixing the number of iterations of $D$ and $G$ can not be applied effectively to a wide variety of time series datasets.
Motivated by the above considerations, we adopt an adaptive training strategy to maintain a balance between $G$ and $D$. Specifically, for each batch, $D$ is updated only if its accuracy in the last batch is no larger than a predefined threshold $\delta \in (0.5, 1.0)$. We found that $\delta=0.75$ can stabilize the training and derive more reasonable results. It should be noted that the threshold $\delta$ is not very sensitive, so another similar value is also permissible.

\begin{algorithm}[h]
\caption{Training TCGAN. Default hyper-parameters: $m=16, n_{epoch}=300, P_{z}=Uniform(-1,1), n_{z}=100, \delta = 0.75, \alpha=0.0002, \beta_{1}=0.5, \beta_{2}=0.9$.}
\label{alg:tsgan}
\begin{flushleft}
\textbf{Input}: An unlabeled time series dataset of size $n_{sample}$ and its distribution is represented as $P_{data}$.\\
\textbf{Hyper-parameters}: The batch size, $m$. The number of passes on data, $n_{epoch}$. The noise prior, $P_{z}$, and the length of a noise vector, $n_{z}$. The accuracy threshold triggering $D$ to update, $\delta$. Adam hyper-parameters include the learning rate $\alpha$, the exponential decay rate for the 1st moment estimates $\beta_{1}$ and the 2nd moment estimates $\beta_{2}$. 
\end{flushleft}
\begin{algorithmic}[1] 
\STATE $n_{batch}=\left \lfloor n_{sample} / m \right \rfloor$
\STATE $acc_{last}=\delta$
\FOR{$e=0; e<n_{epoch}; e++$} 
  \FOR{$j=0; j<n_{batch}; j++$}
    \STATE sample $\{\mathbf{x}^{(i)}\}_{i=1}^{m}$ from $P_{data}$
    \STATE sample $\{\mathbf{z}^{(i)}\}_{i=1}^{m}$ from $P_{z}$
    \IF{$acc_{last} <= \delta$}
        \STATE $Loss_{d} = -\frac{1}{m}\sum_{i=1}^{m}[log(D(\mathbf{x}^{(i)})) + log(1-D(G(\mathbf{z}^{(i)})))]$
        \STATE $\mathbf{\theta_{d}} = Adam(\bigtriangledown_{\mathbf{\theta_{d}}}Loss_{d}, \mathbf{\theta_{d}}, \alpha, \beta_{1}, \beta_{2})$
    \ENDIF
    \STATE $Loss_{g} = -\frac{1}{m}\sum_{i=1}^{m}log(D(G(\mathbf{z}^{(i)})))$
    \STATE $\mathbf{\theta_{g}} = Adam(\bigtriangledown_{\mathbf{\theta_{g}}}Loss_{g}, \mathbf{\theta_{g}}, \alpha, \beta_{1}, \beta_{2})$ 
    \STATE $acc_{last}$ = \textit{Acc}($\{<D(\mathbf{x}^{(i)}), 1>\}_{i=1}^{m}, \{<D(G(\mathbf{z}^{(i)})),0>\}_{i=1}^{m}$)
  \ENDFOR
\ENDFOR
\end{algorithmic}
\end{algorithm}

\subsection{Representation Encoder}

We reuse parts of the discriminator to construct a representation encoder, $E(\mathbf{x};\mathbf{\theta_{e}}): \mathbf{x} \in \mathbb{R}^{n*d} \to \mathbf{v} \in \mathbb{R}^{n_{v}}$, that transforms each time series into a representation vector. 
Specifically, we follow common practices in the DNN community to frame the encoder. First, the encoder feeds the input of real time series to the pretrained discriminator and uses the feature maps from the last Conv layer (i.e., Conv4). Then, a max pooling layer with a pooling size of 2 and a stride of 1 is applied to these feature maps. The final representation vector is derived with a flatten layer. 
We will describe the design details below.

Although the discriminator doesn't see the real labels, we believe it has learned how to distill salient features from raw data by competing against the generator during the adversarial game. In fact, the discriminator is a supervised FCN that has been demonstrated to be powerful in feature learning \cite{lecun2015deep,ismail2019deep}. 
Furthermore, the idea that the discriminator learns from a surrogate binary classification task in a GAN architecture is similar to the approach of using surrogate labels to realize self-supervised learning \cite{doersch2017multi,jenni2018selfsupervised}. In the computer vision domain \cite{creswell2018generative}, the GAN discriminator has been demonstrated to be helpful for a variety of downstream tasks. As shown in Fig.\ref{intro-case}, time series of different classes do become easier to distinguish in the TCGAN transformation space.

For compact representation, we only reuse outputs from one Conv layer, instead of fusing all layers. We opt to use the last Conv layer because it is widely recognized that the deeper a layer is located in a neural network, the more abstract the underlying representation will be \cite{lecun2015deep}. 

The superiority of CNNs in time seres feature learning has been illustrated in recent literature \cite{ismail2019deep}. In fact, convolution is a well-established method for handling sequential signals \cite{mallat1999wavelet}.
Suppose $\mathbf{x} \in \mathbb{R}^{n*d}$ is a time series of length $n$ and $\mathbf{f} \in \mathbb{R}^{w*d}$ is a filter of length $w$. Let $(\mathbf{x}*\mathbf{f})$ denote the result of one-dimensional discrete convolution. A general form of performing convolution at a time stamp $t$ is given by Eq.~\eqref{eq:1dconv} where $b_{t}$ is an independent bias. The filter is shared by all time stamps to extract time-invariant features across the whole time series.
The result of one filter can be viewed as a transformation on the input time series. And thus, one filter detects one type of features. Multiple filters in a Conv layer extract different types of features. 
\begin{equation}
\label{eq:1dconv}
(\mathbf{x}*\mathbf{f})[t] = \sum_{i=0}^{w-1}\mathbf{x}_{t+i} \cdot \mathbf{f}_{w-i} + b_{t}, t \in \{1, 2,..., n\}
\end{equation}

We use a max-pooling layer right after the Conv feature maps to enable the transformations to be invariant to shifts and distortions \cite{jarrett2009best,lecun1990handwritten}. Specifically, for a feature map $\mathbf{h} \in \mathbb{R}^{n*m}$ with scale $(n, m)$, where $n$ denotes the length of the input sequence and $m$ denotes the number of channels, the max-pooling layer applies a sliding window of length $w$ over each channel and results in a new feature map $\mathbf{h'}$. Each element of $\mathbf{h'}$ is given by Eq.~\eqref{eq:maxpool}.
\begin{equation}
\begin{aligned}
\label{eq:maxpool}
\mathbf{h'}_{i,k} = max\{\mathbf{h}_{j,k}\}_{j=\frac{n}{w}(i-1)+1}^{\frac{n}{w}i}; \\ i=\{1,...,n-w+1\}, k=\{1,...,m\}. 
\end{aligned}
\end{equation}

It should be noted that an appropriate normalization could be applied to the final representations, it depends on the downstream models or tasks. We use non-normalized features by default. 

\subsection{Classification and Clustering}
\label{sec:approach:cla&clus}

TCGAN representations can, in principle, be used in many time series mining tasks. We focus on classification and clustering tasks. We will demonstrate in Section \ref{sec:exp} that TCGAN representations are very effective in enabling linear models to achieve superior performance. 
It should be noted that linear models are simple and fast, but they need a nonlinear transformation of the input to process complicated data. For example, Support Vector Machine (SVM) usually applies non-linear kernel functions to make it applicable for nonlinearly separable data spaces. However, it is hard to find a suitable kernel function to improve performance, and the computation of kernels is costly and potentially infeasible. In contrast, TCGAN transformations can be learned automatically and efficiently.

For classification, we consider the following linear classifiers. Linear Support Vector machine for Classification (LSVC) and Logistic Regression(LR) are common in the machine learning community and suitable for very large datasets. SoftMax (SM) is always the last layer of a DNN for classification.  

For clustering, k-means with Euclidean distance (k-means) is the most basic clustering method and assumes inputted instances can be represented as points in an Euclidean space. 

Time series datasets tend to be small, and thus it is hard to split a validation set for careful hyper-parameter tuning. Therefore, we use the default settings for the above models provided in the popular machine learning toolkit scikit-learn, with the exception that SM applies a learning rate of 0.0002 and an epoch of 100. 

\section{Experiments}
\label{sec:exp}

We conduct comprehensive experiments on synthetic and real time series datasets to answer the following questions: 
\begin{enumerate}[1)]
  \item How does TCGAN learn through the adversarial game? (Subsection \ref{sec:gans-tcgan})
  \item Compared with the leading time-series GANs, how does TCGAN perform on generation and classification tasks? (Subsection \ref{sec:gans-comp})
  \item Compared with the superior supervised and unsupervised CNNs, how effective is TCGAN for general time series classification? (Subsection \ref{sec:tcgan-tsc})
  \item How robust is the TCGAN classifier when the labeled data becomes highly limited and/or imbalanced? (Subsection \ref{sec:tcgan-tsc-extreme})
  \item Are the pairwise similarities between time series well preserved in the TCGAN transformation for the clustering task? (Subsection \ref{sec:tcgan-repr-space})
\end{enumerate}
In addition, we report TCGAN's runtime in Subsection~\ref{sec:runtime}. 
Before presenting experimental results, we describe datasets, evaluation metrics, and implementation in Subsection \ref{sec:setup}. 

\subsection{Experiment Setup}
\label{sec:setup}

\subsubsection{Datasets}
We include 85 time series datasets from the UCR repository \cite{ucr15dataset} and synthetic Sines data from \cite{yoon2019time}.

Following UCR repository \cite{ucr15dataset}, we use the default training/test splits. UCR datasets are already $z$-normalized. The datasets vary in type of data, the size of the training set, the size of the test set, the length of the time series and the number of classes.
\begin{itemize}
  \item \cite{bagnall2016great} categorizes the datasets into 7 types: Image Outline (29 datasets), Sensor Readings (16), Motion Capture (14), Spectrograph (7), ElectroCardioGraph (ECG) measurements (7), Electric Devices (6) and Simulated (6). 
  \item The size of the training set is small. The number of training instances ranges from 16 (DiatomSizeReduction dataset) to 8926 (ElectricDevices dataset), with an average of 432. There are 33 datasets with sizes in the range of $[16, 100]$, 36 in the range of $[101, 500]$, and 16 in the range of $[501, 8926]$.
  \item In most cases, the test set is larger than the training set, ranging from 20 (BeetleFly dataset) to 8236(StarLightCurves dataset), with an average of 1164. There are 29 datasets with sizes in the range of $[20, 300]$, 32 in the range of $[301, 1000]$, and 24 in the range of $[1001, 8236]$.
  \item The length of the time series ranges from 24 (ItalyPowerDemand dataset) to 2709 (HandOutlines dataset), with an average of 422. There are 43 datasets with lengths in the range of $[24, 300]$, 25 datasets in the range of $[301, 700]$ and 17 datasets in the range of $[701, 2709]$.
  \item The number of classes ranges from 2 to 60 (ShapesAll dataset), with an average of 7. There are 71 datasets with numbers in the range of $[2, 10]$, 8 datasets in the range of $[11, 30]$, and 6 datasets in the rang of $[31, 60]$.
\end{itemize}

Following \cite{yoon2019time}, we synthesize univariate and multivariate sinusoids with different frequencies and phases (Sines) for experiments. For each variable, sinusoidal sequences are sampled at different frequencies $\theta \sim Uniform(0,1)$ and phases $\theta \sim Uniform(-\pi,\pi)$, i.e., $x = sin(2\pi\eta t + \theta)$. We consider two types of Sines data: 
\begin{itemize}
  \item SinesD1L100: includes 1-dimensional Sine time series with a length of 100.
  \item SinesD5L24: includes 5-dimensional Sine time series with a length of 24.  
\end{itemize}
For each repeated experiment, we synthesized 10,000 samples to construct a dataset.

\subsubsection{Evaluation metrics}

Following existing works \cite{esteban2017real,creswell2018generative,yoon2019time,xu2020cot}, we evaluate time-series GANs by assessing the quality of the generated data, i.e., fidelity and diversity. We also consider the efficiency of GANs. 
To evaluate the fidelity, we visualize some generated samples and apply two metrics to quantitatively measure the similarity between real data and generated data. Specifically, for a generated dataset of the same size as the training dataset, we calculate the Nearest Neighbor Distance (NND) and Maximum Mean Discrepancy (MMD) metrics \cite{esteban2017real,gretton2007kernel}.
Usually, a smaller value for both metrics signifies better GAN performance, generally the generator. Assume the real dataset is $\{\mathbf{x}^{i}\}_{i=1}^{N}$ and the generated dataset is $\{\mathbf{\hat{x}}^{i}\}_{i=1}^{N}$. MMD is defined as Eq.~\ref{eq:mmd}, where $K(., .)$ is the L2-distance-based radial basis function. 
\begin{equation}
\begin{aligned}
\label{eq:mmd}
\hat{MMD}^{2}=\frac{1}{N(N-1)}\sum_{i=1}^{N}\sum_{j\neq i}^{N}K(\mathbf{x}^{i}, \mathbf{x}^{j}) \\
+\frac{1}{N(N-1)}\sum_{i=1}^{N}\sum_{j\neq i}^{N}K(\mathbf{\hat{x}}^{i}, \mathbf{\hat{x}}^{j}) \\
- \frac{2}{NN}\sum_{i=1}^{N}\sum_{j=1}^{N}K(\mathbf{x}^{i}, \mathbf{\hat{x}}^{j})
\end{aligned}
\end{equation}
As defined in Eq.~\ref{eq:nnd}, NND is estimated by first calculating the distance of each generated sample from the nearest neighbor in the real dataset and then aggregating them all.
\begin{equation}
\label{eq:nnd}
NND = \frac{1}{n}\sum_{i=1}^{N}min_{j=1,...,N}\left \| \mathbf{\hat{x}}^{i} - \mathbf{x}^{j} \right \|^{2}
\end{equation}
To assess the diversity of generated data, we utilize t-SNE \cite{maaten2008visualizing} to visualize how closely the distribution of the generated data matches that of the original data in 2-dimensional space. Additionally, we measure the efficiency of the GANs by calculating their training and inference time.

We specifically evaluate the usefulness of GANs for time series classification and clustering. 
To measure classification performance, we use the common metric, accuracy, which counts the proportion of samples that are correctly classified. For imbalanced classification scenarios, we use the weighted F1 score. A weighted F1 score is the average of binary metrics (F1 scores) weighted by the number of true instances for each class.
Following \cite{lei2019similarity}, we apply normalized mutual information (NMI) on the fused training and test splits, and set the number of classes ($C$) as the target number of clusters. Eq.~\ref{eq:nmi} defines the NMI, where $\mathbf{U}$ and $\mathbf{V}$ are two label assignments of the same $N$ samples in $C$ classes, $|\mathbf{U}_{i}|$ and $|\mathbf{V}_{j}|$ are the number of samples in cluster $\mathbf{U}_{i}$ and $\mathbf{V}_{j}$. $|\mathbf{U}_{i} \cap \mathbf{V}_{j}|$ denotes the number of samples that belong to the intersection of sets $\mathbf{U}_{i}$ and $\mathbf{V}_{j}$. An NMI value close to 1 indicates high-quality clustering.
\begin{equation}
\label{eq:nmi}
NMI = \frac{\sum_{i=1}^{C}\sum_{i=1}^{C}|\mathbf{U}_{i}\cap \mathbf{V}_{j}|log(\frac{N|\mathbf{U}_{i}\cap \mathbf{V}_{j}|}{|\mathbf{U}_{i}| |\mathbf{V}_{j}|})}{\sqrt{(\sum_{i=1}^{C}|\mathbf{U}_{i}|log\frac{|\mathbf{U}_{i}|}{N})(\sum_{j=1}^{C}|\mathbf{V}_{j}|log\frac{|\mathbf{V}_{j}|}{N})}}
\end{equation}

We compare two methods on multiple datasets in pairs and calculate significance using the Wilcoxon signed rank test (WSRT) with Holm correction \cite{wilcoxon1992individual,garcia2008extension}. We also put multiple competitors in a Critical Difference (CD) diagram \cite{demvsar2006statistical} for compact comparison, where a thick horizontal line represents a group of approaches (a clique) that are not significantly different.

\subsubsection{Implementation}

We implemented our codebase in Python-3.6.13. The deep learning models were implemented with TensorFlow-2.3.0 and other machine learning models, e.g., conventional classification and clustering models, were implemented with scikit-learn-0.24.2. 
We ran our experiments on a server featuring 16 2.50GHz Intel(R) Xeon(R) E5-2682 CPU cores and a NVIDIA Tesla P100 GPU.
If not specified, we used a uniform set of hyper-parameters on all datasets. The settings of TCGAN were presented in Alg.\ref{alg:tsgan} and Section \ref{sec:approach}. For benchmarked methods, we followed their original settings.
All unsupervised procedures were trained on the fused training and test splits, and all supervised procedures were applied to the training split only. 
For methods with randomness, their results are reported using an average of five runs with different random seeds.

\subsection{Training of TCGANs} 
\label{sec:gans-tcgan}

We trained TCGAN using Alg.\ref{alg:tsgan} for each dataset without any label information. 

\begin{figure*}[h]
\centering
\includegraphics[width=0.8\linewidth]{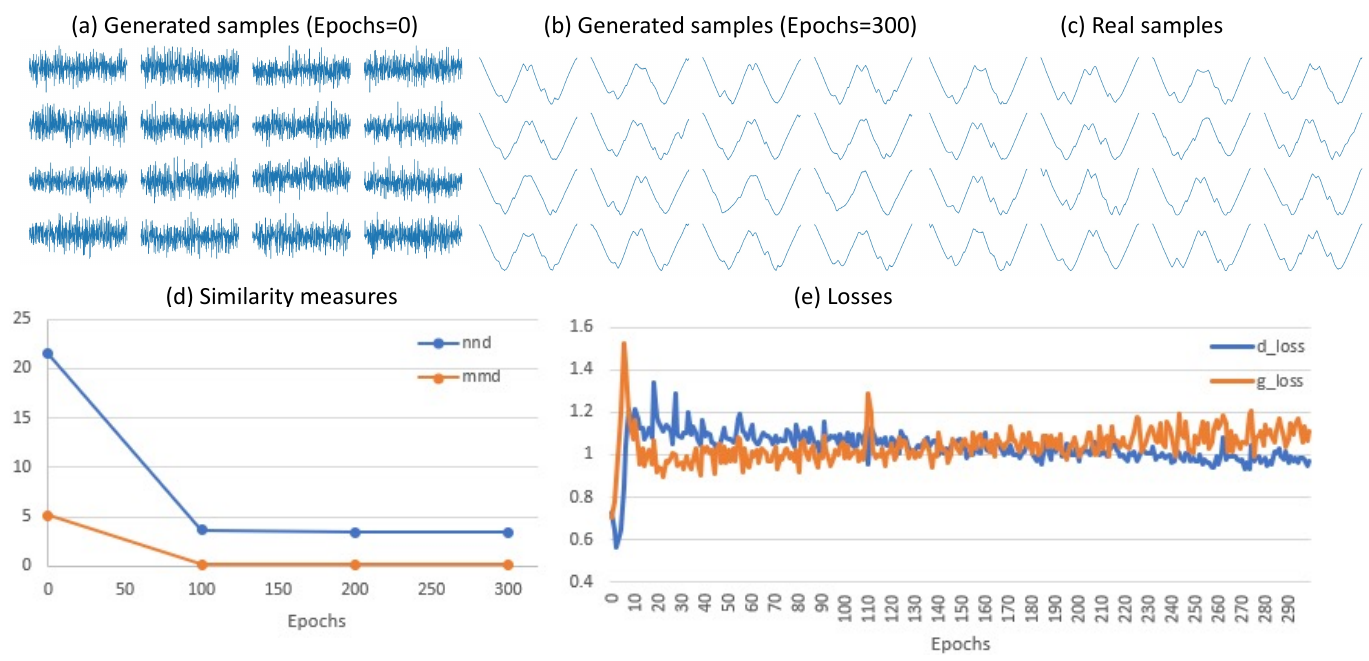}
\caption{Training and Evaluation of TCGAN on the FISH dataset.}
\label{fig:gan-show-uts}
\end{figure*}

Fig.~\ref{fig:gan-show-uts} shows the training and evaluation results of TCGAN on the FISH dataset. 
The results indicate that both the generator ($G$) and discriminator ($D$) of TCGAN do learn from the adversarial game and converge to a stable level. 
Specifically, Fig.~\ref{fig:gan-show-uts}(e) shows the the competition between $G$ and $D$. In the early stage, the loss of $D$ (d\_loss) drops sharply to a low level. This phenomenon signifies that $D$ can easily learn to solve the discrimination task. Meanwhile, the increasing loss of $G$ (g\_loss) suggests that $G$ is receiving a stronger signal from $D$ to improve itself. As $G$ begins to generate more realistic samples, the discrimination task of $D$ becomes harder, and thus d\_loss turns to increase. Finally, $G$ and $D$ reach an adversarial equilibrium. 
Similarly, in Fig.~\ref{fig:gan-show-uts}(d), both NND and MMD, which measure the similarity between generated and real samples, also converge to a relatively low level.  
Visually inspecting the generated time series, reveal noisy signals in the beginning (Fig.~\ref{fig:gan-show-uts} (a)) and appear more realistic in the end (Fig.~\ref{fig:gan-show-uts} (b)) becoming nearly identical with the ground truth (Fig.~\ref{fig:gan-show-uts} (c)).

\begin{figure*}[h]
\centering
\includegraphics[width=0.8\linewidth]{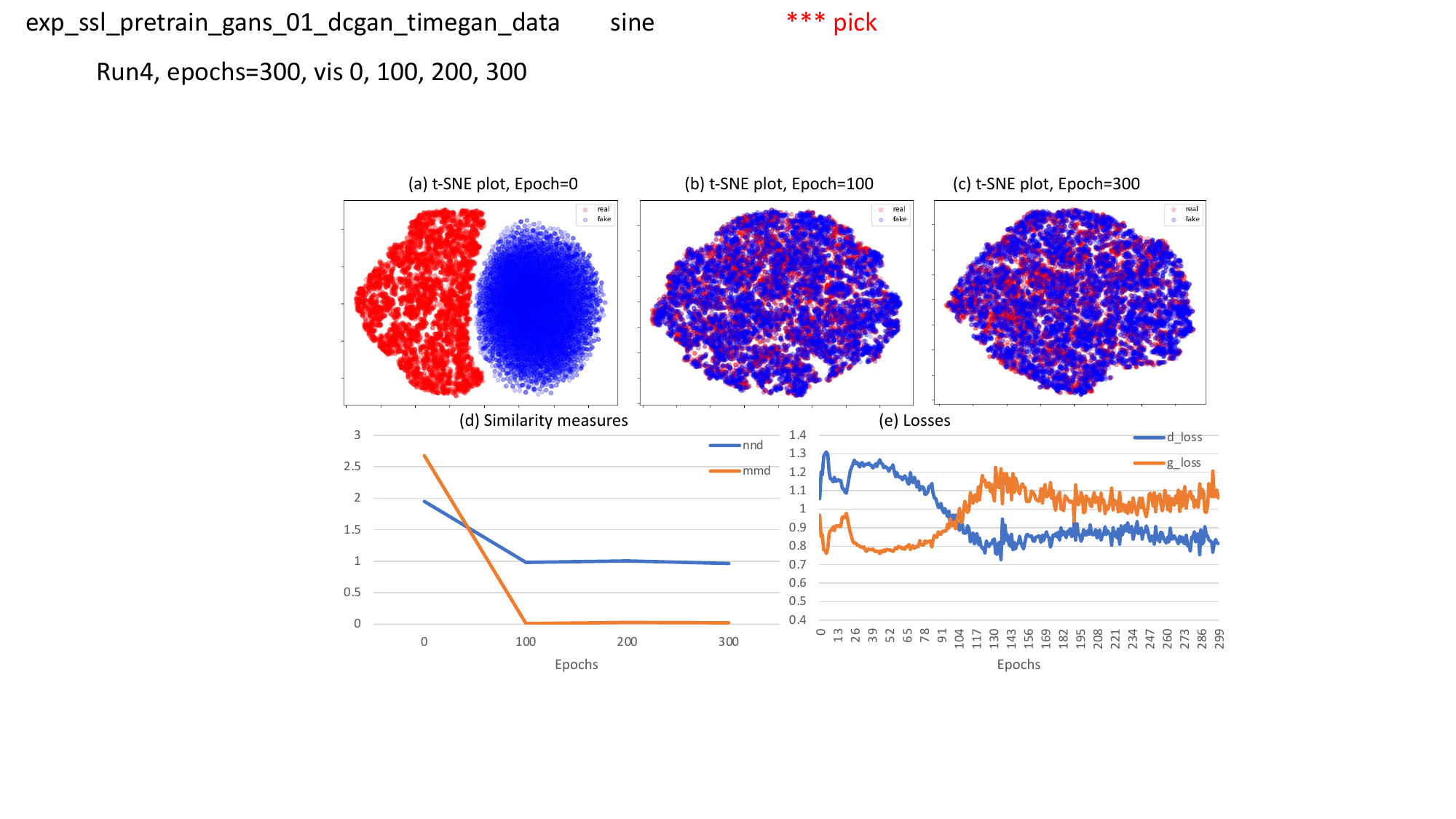}
\caption{Training and Evaluation of TCGAN on the SinesD5L24 dataset.}
\label{fig:gan-show-mts}
\end{figure*}

Fig.~\ref{fig:gan-show-mts} shows another case on the SinesD5L24 dataset. 
The results further demonstrate that TCGAN effectively learn during the adversarial game between $G$ and $D$. Specifically, Fig.~\ref{fig:gan-show-mts}(a)-(c) are t-SNE visualizations of real data (red dots) and generated data (blue dots) at different epochs during the learning process. The real samples and generated samples move closer and closer until they perfectly overlap, indicating that the generated samples distributively cover the real data. Meanwhile, similarity measures (NND and MMD) in Fig.~\ref{fig:gan-show-mts}(d) consistently converge to a low level.
Fig.~\ref{fig:gan-show-mts}(e) illustrates the healthy competition between $G$ and $D$, they alternate between strength and weakness before reaching an adversarial equilibrium.

\subsection{Comparisons of Time-series GANs}
\label{sec:gans-comp}

In the following two subsections, we compare different GANs in terms of their generation performance against synthetic Sine datasets, as well as their classification performance on UCR datasets.
We used the publicly available source code to implement TimeGAN \cite{yoon2019time} \footnote{{TimeGAN source code: \url{https://github.com/jsyoon0823/TimeGAN}}} and CotGAN \cite{xu2020cot} \footnote{{CotGAN source code: \url{https://github.com/tianlinxu312/cot-gan}}}. 
We followed each model's default settings with the exception of using the same number of epochs (i.e., 300) for the adversarial training. It should be noted that, prior to adversarial training, TimeGAN has to pretrain its auto-encoder component over 100 epochs and its supervised teacher-forcing component over 100 epochs. Therefore, TimeGAN actually undergoes 500 epochs in total.
For more details about the competitors, please refer to Section~\ref{sec:rel-gan}.

\subsubsection{Time-series GANs for Generation}

\begin{table*}[h]
\caption{Quantitative performance of Time-series GANs on Sines data. The mean and standard deviation (in bracket) are presented for each entry. \textbf{Bold} indicates best performance.}
\label{tab:gen}
\centering
\begin{tabular}{l|cccc}  
\toprule
        & MMD    &NND   & InferTime(s)$^a$  & TrainTime(s)$^b$\\
\midrule
\multicolumn{5}{c}{SinesD1L100 Dataset}   \\
\hline
TimeGAN & 0.0647(0.0848)   &1.4189(1.5016)  & 1.1682(0.1235)   & 63.7668(1.8127) \\
CotGAN & 0.0222(0.0116)    &1.4330(0.2610)   & 0.8183(0.0150)   & 57.6786(2.1978) \\ 
TCGAN & \textbf{0.0012(0.0008)} & \textbf{0.1112(0.0262)}   & \textbf{0.0586(0.0040)}  & \textbf{0.8225(0.0181)} \\ 
\midrule
\midrule
\multicolumn{5}{c}{SinesD5L24 Dataset}   \\
\hline
TimeGAN & 0.0244(0.0131)   &1.2581(0.0740)  & 0.4829(0.0765)   & 32.6466(1.3590) \\
CotGAN & 0.0054(0.0005)    &1.4568(0.0317)   & 1.4432(0.1281)   & 103.8826(0.5975) \\ 
TCGAN & \textbf{0.0050(0.0012)} & \textbf{0.9394(0.0067)}   & \textbf{0.0469(0.0011)}  & \textbf{0.8593(0.0151)} \\ 
\hline
\multicolumn{5}{l}{\footnotesize{$^a$ The time (seconds) for generating a dataset with 10,000 samples.} }\\
\multicolumn{5}{l}{\footnotesize{$^b$ The time (seconds) for 1 epoch adversarial/joint training.} } \\
\bottomrule
\end{tabular}
\end{table*}

\begin{figure*}[h]
\centering
\includegraphics[width=0.8\linewidth]{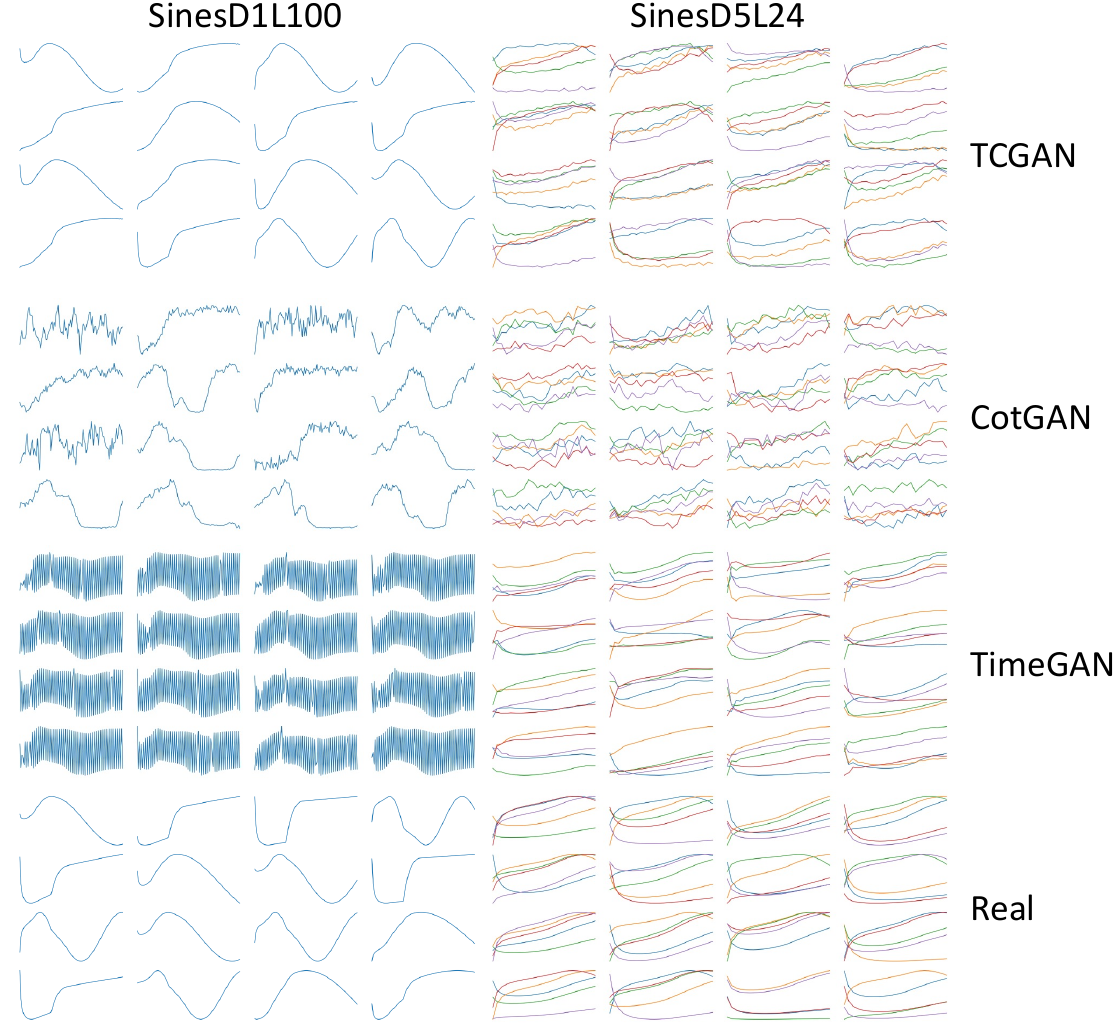}
\caption{Samples of SinesD1L100 ($1^{st}$ column) and SinesD5L24 ($2^{nd}$ column) obtained by different methods. Each of the top three rows presents the generated samples for each GAN. Bottom row corresponds to the real samples. }
\label{fig:exp-tsgan-samples}
\end{figure*}

\begin{figure*}[h]
\centering
\includegraphics[width=0.8\linewidth]{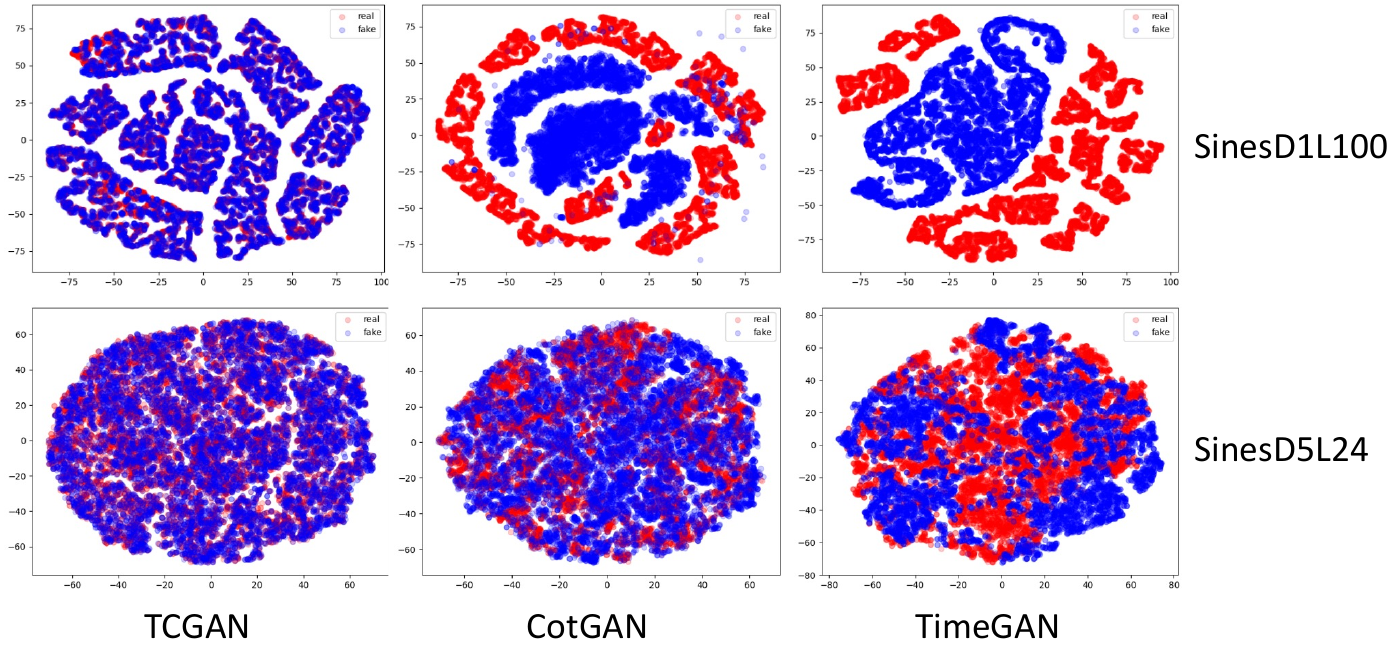}
\caption{t-SNE visualization on SinesD1L100 ($1^{st}$ row) and SinesD5L24 ($2^{nd}$ row). Each column corresponds to one model. Red denotes original data, and blue denotes generated data.}
\label{fig:exp-tsgan-tsne}
\end{figure*}

We compare TCGAN with TimeGAN and CotGAN on the SinesD1L100 and SinesD5L24 datasets. Following \cite{yoon2019time}, we use a batch size of 128. 
After training, each GAN is used in test mode to generate a set of time series of the same size as the training set for evaluation.
In Tab.\ref{tab:gen} reports the mean and standard deviation of quantitative evaluation statistics over 5 random runs. We visualize samples from one single run in Fig.\ref{fig:exp-tsgan-samples} and Fig.\ref{fig:exp-tsgan-tsne}. 

First, we evaluate the fidelity of generated samples of different GANs. According to the MMD and NND in Tab.\ref{tab:gen}, TCGAN consistently produces the best values. The results indicate that the fake samples generated by TCGAN are closer to the real samples compared with TimeGAN and CotGAN. 
Fig.\ref{fig:exp-tsgan-samples} presents several generated and real samples. For the SinesD1L100 dataset, TCGAN samples appear almost identical to the real samples, while CotGAN only simulates rough contours, and TimeGAN generates flat signals significantly different from the real. For the SinesD5L24 dataset, TimeGAN samples are the closest to the real, TCGAN samples are very similar to the real but display some weak fluctuations, and CotGAN samples are significantly different from the real.

Second, t-SNE visualizations are used in Fig.\ref{fig:exp-tsgan-tsne} to assess the diversity of the generated samples.
We observe that the generated samples of TCGAN show dramatically better overlap with the original samples than other GANs. In fact, in the first column, the blue (generated) samples and the red (original) samples are almost perfectly in sync. In contrast, CotGAN and TimeGAN simulate short time series (SinesD5L24 dataset) to some extent and almost fail to capture long time series (SinesD1L100 dataset).

Finally, we investigate the efficiency of different GANs. 
The results in Tab.\ref{tab:gen} show that TCGAN is definitively the fastest model for both training and inference.
On SinesD1L100 dataset, TCGAN is $14 \times$ faster in inference and $70 \times$ faster in training than the runner-up CotGAN. 
On SinesD5L24 dataset, TCGAN is $ 10 \times$ faster in inference and $38 \times$ faster in training than the runner-up TimeGAN. 
Furthermore, TCGAN is stable enough to achieve similar performance in both cases. In contrast, TimeGAN is slow to process long time series (SinesD1L100 datasets), and CotGAN is slow to process multi-dimensional time series (SinesD5L24 dataset). 

In summary, the above results demonstrate that TCGAN efficiently generates high-quality time series by avoiding recurrence and stacking convolution layers to model entire time series. 

\subsubsection{Time-series GANs for Classification}

We evaluate the effectiveness of GANs for time series classification, using partial UCR datasets. That is 19 datasets of length up to 100, because CotGAN and TimeGAN are really time-consuming.
We train GANs on the fused training and test splits in the absence of any label information, and then reuse partial pretrained modules to encode raw time series for the same off-the-shelf linear classifier (i.e., a softmax layer). The resulting classifiers are trained on the training split for 100 epochs. We use a small batch size (i.e., 16) because most UCR datasets contain limited samples.

The details for constructing GAN-based classifiers are as follows.
We reuse TCGAN's discriminator (TCGAN) as the representation encoder.
Similarly, TimeGAN has three modules that could be reused: the encoder in the auto-encoder component (TimeGAN-E), the supervised teacher-forcing component (TimeGAN-S) and the discriminator (TimeGAN-D).
CotGAN has two discriminators, $D_{h}$ and $D_{M}$ jointly approximate the optimization objective. Therefore, we reuse $D_{h}$ and $D_{M}$ to form CotGAN-Dh and CotGAN-DM, respectively.
For each of the above representation encoders, we flatten the outputs from the last hidden layer and feed them into a softmax layer to conduct classification. 

Tab.\ref{tab:gen-cls} illustrates the results. For TimeGAN and CotGAN that derive multiple encoders or classifiers, we report the best accuracy of all options. 
The results show that TCGAN significantly outperforms the competitors and wins 15 out of 19 datasets. We attribute this success to the stacking convolution layers in TCGAN that can model contextual and hierarchical features inherent in raw time series. In contrast, CotGAN and TimeGAN depend on auto-regressive modules that prefer to attend to values in the recent past, and the accumulation process may have the vanishing issue and accumulation errors. We notice that the authors of TimeGAN usually use a window size of 24 to slice the time series. 
We also observe that TimeGAN outperforms CotGAN on almost all datasets (18 datasets). However, the results presented in the previous section (see Tab.\ref{tab:gen}) show the generation performance of CotGAN is close to or slightly better than TimeGAN. This phenomenon suggests that pursuing strong performance on the generation task, which is the focus of existing time-series GANs, does not definitely bring benefits to the discrimination task. 

\begin{table*}[h]
\caption{GANs for time series classification on 19 UCR datasets of length up to 100. Each entry corresponds to the mean and standard deviation (in bracket) of accuracies. \textbf{Bold} indicates best performance.}
\label{tab:gen-cls}
\centering
\begin{tabular}{l|lll} 
\toprule
Dataset  & CotGAN  & TimeGAN & TCGAN \\
\midrule
DistalPhalanxOutlineAgeGroup & 0.8066 (0.0011) &0.7925 (0.0045) & \textbf{0.8265 (0.0471)} \\
DistalPhalanxOutlineCorrect  & 0.6137 (0.0022) &0.6857 (0.0189) & \textbf{0.8183 (0.0136)} \\
DistalPhalanxTW & 0.7369 (0.0011) &0.7797 (0.0027) & \textbf{0.79 (0.0085)} \\
ECG200 & 0.8132 (0.0066) &0.8196 (0.0068) & \textbf{0.926 (0.0089)} \\
ElectricDevices & 0.2424 (0.0363) &0.6072 (0.0030) & \textbf{0.628 (0.027)} \\
ItalyPowerDemand & 0.7909 (0.0348) &0.9630 (0.0015) & \textbf{0.9679 (0.0036)} \\
MedicalImages & 0.5175 (0.0157) &0.6304 (0.0132) & \textbf{0.7676 (0.0131)} \\
MiddlePhalanxOutlineAgeGroup & 0.2845 (0.0024) &\textbf{0.7919 (0.0062)} &0.779 (0.0118) \\
MiddlePhalanxOutlineCorrect & 0.5279 (0.0013) &0.5345 (0.0039) &\textbf{0.716 (0.0999)} \\
MiddlePhalanxTW & 0.4566 (0.0641) &\textbf{0.6418 (0.0021)} &0.6226 (0.012) \\
MoteStrain & 0.5428 (0.1236) &\textbf{0.8711 (0.0033)} &0.8246 (0.0114) \\
PhalangesOutlinesCorrect& 0.6689 (0.0143) &0.6924 (0.0067) &\textbf{0.7953 (0.0108)} \\
ProximalPhalanxOutlineAgeGroup & 0.5065 (0.0132) &\textbf{0.8500 (0.0032)} &0.8098 (0.0329) \\
ProximalPhalanxOutlineCorrect & 0.6797 (0.0) &0.8080 (0.0137) &\textbf{0.8577 (0.011)} \\
ProximalPhalanxTW& 0.6721 (0.1225) &0.8008 (0.0018) &\textbf{0.805 (0.0116)} \\
SonyAIBORobotSurface & 0.6202 (0.0117) &0.6848 (0.0055) &\textbf{0.8722 (0.0398)} \\
SonyAIBORobotSurfaceII & 0.7835 (0.0104) &0.8263 (0.0049) &\textbf{0.9158 (0.0095)} \\
synthetic\_control  & 0.6411 (0.0079) &0.8943 (0.0114) &\textbf{0.9913 (0.0069)} \\
TwoLeadECG & 0.699 (0.0633) &0.9476 (0.0025) &\textbf{0.9867 (0.0079)} \\
\midrule
Win & 0 & 4 & 15 \\
\bottomrule
\end{tabular}
\end{table*}

\subsection{TCGAN Representations for Classification}
\label{sec:tcgan-tsc}

As described in section \ref{sec:approach:cla&clus}, we feed TCGAN representations to linear classifiers for time series classification. Using Linear Support Vector Machine for classification (LSVC), Logistic Regression (LR) and SoftMax layer (SM), we have \textbf{LSVC-TCGAN}, \textbf{LR-TCGAN} and \textbf{SM-TCGAN} classifiers, respectively.  

For fair comparison, we use TCGAN as the backbone to construct the following CNN classifiers:  
\begin{itemize}
  \item \textbf{Supervised CNNs}. We replace the sigmoid output layer of TCGAN discriminator with a softmax layer to construct a standard FCN for supervised multi-class classification. We train the resulting model from scratch (\textbf{TCGAN-D}) or only randomly initialize it (\textbf{TCGAN-D-R}). We also include the \textbf{FCN} proposed in \cite{ismail2019deep}. 
  \item \textbf{Unsupervised CNNs}. GANs are innately similar to AutoEncoders (AEs) that have been widely used in unsupervised learning \cite{kingma2019introduction}. We reform TCGAN to construct an AE structure where the discriminator and generator of TCGAN are used as the encoder and decoder of AE. Similarly, we froze the pretrained AE encoder and connect it with a SoftMax layer for classification (\textbf{SM-AE}) as the same as SM-TCGAN. We also consider advanced variants of AE: denoising AE (DAE) and Variational AE (VAE). Similar to  SM-AE, we build \textbf{SM-DAE} and \textbf{SM-VAE} classifiers. In the case of VAE, we reuse the layer before the middle bottleneck layer because we have the same observation as \cite{zhang2019adversarial} that the latent representations output from the bottleneck layer are randomly sampled from a multivariate Gaussian distribution and thus are inappropriate for classification. In fact, the resulting VAE corresponds to the VAE++ proposed in \cite{zhang2019adversarial}. 
\end{itemize}

We also include recent SOTA representation learning methods for time series classification. 
\begin{itemize}
  \item 
  \cite{eldele2022self} extends Time-Series representation learning framework via Temporal and Contextual Contrasting (\textbf{TS-TCC}) to propose Class-Aware TS-TCC (\textbf{CA-TCC}) for semi-supervised time series classification. TS-TCC and CA-TCC are CNNs and use an unsupervised-pretrained encoder for a linear classifier. 
  Specifically, TS-TCC learns representations from unlabeled data with contrastive learning on time-series-specific weak and strong augmentations. TS-TCC has a temporal contrasting module to capture temporal relationships and a contextual contrasting module to capture discriminative representations. The self-supervised pretrained TS-TCC is then used as a representation encoder to collaborate with a linear classifier (i.e., a softmax layer) for time series classification. The resulting classifier has two settings: (1) \textbf{TS-TCC-L(inear)} frozen the encoder and (2) \textbf{TS-TCC-T(une)} fine-tune the encoder in the supervised training process. 
  Based on TS-TCC, CA-TCC adds two phases to utilize labeled data in the supervised learning process. CA-TCC deploys TS-TCC-T, which has been fine-tuned on (limited) real-labeled data, to generate pseudo labels for the entire unlabeled dataset. Then, the unsupervised contextual contrastive module of TS-TCC is replaced with a supervised contextual contrasting module in the CA-TCC to ingest the pseudo labels. 
  We use the open-source code \footnote{TS-TCC and CA-TCC: \url{https://github.com/emadeldeen24/CA-TCC}} to reproduce TS-TCC-L, TS-TCC-T, and CA-TCC. 
  \item 
  TS2Vec is an unsupervised representation learning framework for time series that has demonstrated superior performance in the classification task \cite{yue2021ts2vec}. It employs a Temporal Convolutional Network (TCN) as backbone, incorporating dilated convolutions for time series forecasting. 
  TS2vec performs contrastive learning in a hierarchical manner, considering augmented context views to capture multi-scale features of time series at both the timestamp and the instance levels. Additionally, a random cropping strategy is adopted to generate new contexts for learning position-agnostic representations. 
  To utilize the pretrained representations for classification, the authors employ a non-linear classifier, a SVM with RBF kernel, and selected the penalty C by grid search. We refer to this classifier as \textbf{SVM-TS2Vec}. 
  For fair comparison, we also connect the pretrained TS2Vec with the linear classifiers SM, LSVC and LR and derive \textbf{SM-TS2Vec}, \textbf{LSVC-TS2Vec} and \textbf{LR-TS2Vec}, respectively. 
  We use the open-source code \footnote{TS2Vec: \url{https://github.com/yuezhihan/ts2vec}} to reproduce the TS2Vec and its associated classifiers.
\end{itemize}

\begin{figure*}[h]
\centering
\includegraphics[width=0.8\linewidth]{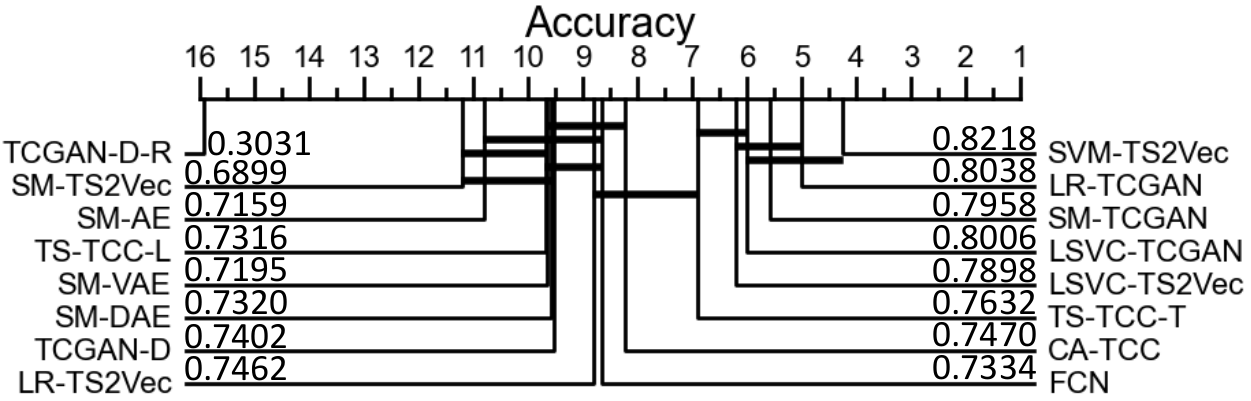}
\caption{Critical difference diagram for CNN classifiers on 85 UCR datasets. The float numbers are the average accuracies.}
\label{fig:clf-fredman-nn}
\end{figure*}

We run experiments on 85 UCR datasets for five repetitions. For a compact comparison, we put the results in a critical difference diagram (Fig.~\ref{fig:clf-fredman-nn}) and have the following observations.
\begin{itemize}
  \item TCGAN classifiers consistently rank in the top group. The results manifest that TCGAN can learn useful representations in an unsupervised manner to boost simple classifiers to achieve superior performance. 
  \item TCGAN-D-R is absolutely the worst model. Therefore, the superior performance of the above TCGAN classifiers does not come from random convolutions.
  \item SM-TCGAN outperforms TCGAN-D by a large margin. This result indicates that, in comparison with training a supervised network from scratch, TCGAN can improve classification performance by making use of additional unlabeled data (samples in test split without label information) for unsupervised pretraining. In addition, TCGAN-D is close to the FCN that verifies the correctness of SM-TCGAN and other variants.
  \item The unsupervised networks (SM-AE, SM-DAE and SM-VAE) are inferior to the supervised networks (FCN and TCGAN-D). This phenomenon aligns with the conclusions in \cite{ismail2019deep}. Again, the results demonstrate the superiority of TCGAN by significantly outperforming peer supervised networks via the same unsupervised learning schema as AEs.
  \item 
  TS-TCC-L is close to the AE variants, while TS-TCC-T is competitive with LSVC-TSGAN. This phenomenon suggests that the representations learned in the pretrained TS-TCC are not well aligned with the real labels, and thus fine-tuning is required to break the gap for better performance. In comparison, the frozen pretrained TCGAN works well with the linear classifiers. 
  CA-TCC is better than TS-TCC-L with the benefit of fine-tuning, while it is worse than TS-TCC-T, indicating that pseudo labels can have negative effects. In fact, CA-TCC is designed for semi-supervised learning with limited real-labeled data. We will have a further discussion in Section~\ref{sec:tcgan-tsc-extreme-ssl}.
  \item 
  SVM-TS2Vec achieves the highest accuracy, confirming the conclusions drawn in the original paper that emphasize the benefits of fine-grained and multi-scale features. The other models do not encapsulate representations in different levels of granularity like TS2Vec.
  However, we observed that the pretrained TS2Vec representations are not well-suited for linear classifiers. LSVC-TS2Vec, LR-TS2Vec and SM-TS2Vec perform worse than the TCGAN classifiers.  SM-TS2Vec is even the worst classifier except the randomly initialized model TCGAN-D-R. This observation suggests that achieving superior performance with TS2Vec representations requires a strong classifier and careful fine-tuning process.
  Furthermore, in Section~\ref{sec:tcgan-tsc-extreme-ssl}, we will demonstrate that TS2Vec is generally less effective than TCGAN in the semi-supervised scenario.
\end{itemize}
The above results demonstrate that TCGAN does automatically learn a good representation of raw data that makes the classification task easier. 

To investigate the high variance problem mentioned in Fig.~\ref{intro-dnn}, we compare SM-TCGAN with FCN on 18 UCR datasets over which FCNs have standard deviations greater than 0.1. TCGAN wins 15 datasets, and all results have dramatically smaller standard deviations. In average, SM-TCGAN ($0.8188 \pm 0.0259$) is significantly more accurate and stable than FCN ($ 0.6440 \pm 0.1657$). The result suggests that TCGAN shows promise in alleviating the high variance problem widely prevalent in supervised networks.
We attribute this improvement to the unsupervised pretraining network in SM-TCGAN. In fact, joint training the layers of a supervised DNN is difficult, and the network can easily converge to a bad local minimum with an inappropriate initialization. As a result, the performance tends to be unstable across multiple runs with different random initializations, particularly for small data.
Existing works \cite{erhan2010does} have demonstrated that unsupervised pretraining has a regularization effect and can guide the learning towards basins of attraction of minima that support better generalization from the training dataset. 

\begin{table*}[h]
\caption{Mean and standard deviation (Std) of classification accuracies. On the selected 18 UCR datasets, FCNs have standard deviations greater than 0.1.}
\label{tab:clas}
\centering
\begin{tabular}{l|ll|ll} 
\toprule
 & \multicolumn{2}{c}{Mean}  & \multicolumn{2}{c}{Std} \\
Dataset  & FCN  & TCGAN   & FCN  & TCGAN \\
\midrule
ArrowHead  & 0.672 & \textbf{0.8514} & 0.1973 & \textbf{0.0114}\\
Beef  & 0.4733 & \textbf{0.84} & 0.1321 & \textbf{0.0279}\\
BirdChicken  & \textbf{0.88} & 0.69 & 0.1151 & \textbf{0.0894}\\
Car  & 0.6667 & \textbf{0.88} & 0.172 & \textbf{0.0139}\\
CBF  & 0.9271 & \textbf{0.9964} & 0.1475 & \textbf{0.0009}\\
ChlorineConcentration  & 0.4661 & \textbf{0.8837} & 0.1646 & \textbf{0.0067}\\
CinC\_ECG\_torso  & 0.6351 & \textbf{0.9722} & 0.1181 & \textbf{0.0236}\\
Cricket\_Y  & 0.6764 & \textbf{0.7287} & 0.11 & \textbf{0.0201}\\
DistalPhalanxOutlineCorrect  & 0.6757 & \textbf{0.8183} & 0.1311 & \textbf{0.0136}\\
Earthquakes  & 0.5193 & \textbf{0.7752} & 0.2256 & \textbf{0.0135}\\
FaceFour  & 0.7523 & \textbf{0.9091} & 0.2582 & \textbf{0.018}\\
Gun\_Point  & 0.9253 & \textbf{0.9707} & 0.1081 & \textbf{0.0174}\\
MALLAT  & 0.6641 & \textbf{0.9431} & 0.2154 & \textbf{0.0118}\\
Meat  & 0.43 & \textbf{0.8967} & 0.166 & \textbf{0.0075}\\
MiddlePhalanxOutlineCorrect  & \textbf{0.74} & 0.716 & 0.1035 & \textbf{0.0999}\\
OliveOil  & 0.2133 & \textbf{0.5067} & 0.107 & \textbf{0.0596}\\
OSULeaf  & \textbf{0.7802} & 0.6769 & 0.3102 & \textbf{0.0125}\\
SmallKitchenAppliances  & 0.4949 & \textbf{0.6837} & 0.2011 & \textbf{0.0178}\\
\bottomrule
Avg & 0.6440 & \textbf{0.8188} & 0.1657 & \textbf{0.0259} \\
\bottomrule
\end{tabular}
\end{table*}

\subsection{TCGAN Representations for Classification in Extreme Situations}
\label{sec:tcgan-tsc-extreme}

\subsubsection{TCGAN Representations for Classification in Absence of Labeled Data}
\label{sec:tcgan-tsc-extreme-ssl}

\begin{figure*}[h]
  \centering
  \includegraphics[width=0.8\linewidth]{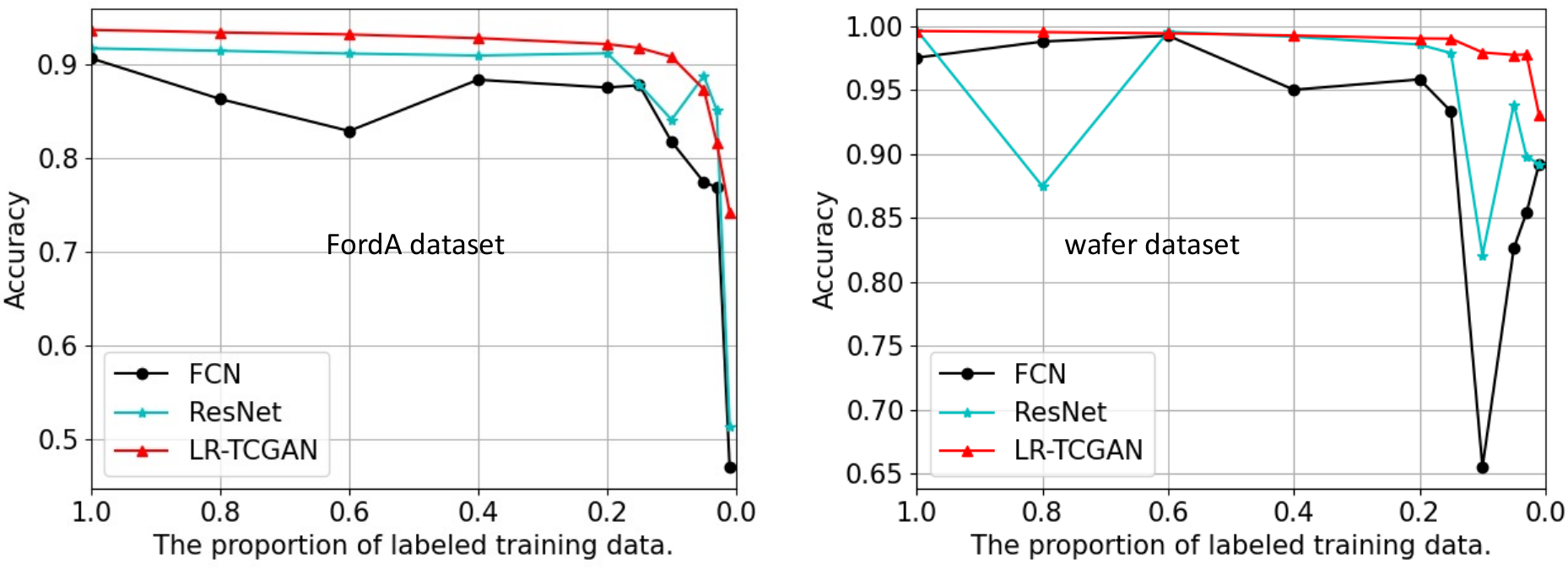}
  \caption{Classification accuracy varies with respect to the size of labeled training dataset.}
  \label{fig:semi-supervised}
\end{figure*}

\begin{table*}[h]
\caption{Classification accuracies of semi-supervised methods with 40\% of labeled training set. \textbf{Bold} indicates best performance. }
\label{tab:ssl40}
\centering
\begin{tabular}{l|ccccc} 
\toprule
Dataset & TCGAN   & CA-TCC & SemiTime & SVM-TS2Vec   & LSVC-TS2Vec \\
\bottomrule
ChlorineConcentration & \textbf{0.7314} & 0.5369 & 0.6242 & 0.5988 & 0.566 \\
Cricket\_X & 0.6251 & 0.5687 & 0.5271 & \textbf{0.6862} & 0.6785 \\
DistalPhalanxOutlineCorrect & \textbf{0.8103} & 0.7453 & 0.7433 & 0.7406 & 0.7399 \\
ElectricDevices & 0.6354 & 0.6752 & 0.306 & \textbf{0.6988} & 0.6743 \\
FordA & 0.9008 & 0.9117 & 0.8905 & \textbf{0.9315} & 0.9182 \\
FordB & \textbf{0.8958} & 0.895 & 0.8776 & 0.7933 & 0.7775 \\
HandOutlines & 0.8554 & 0.8354 & 0.6544 & 0.9103 & \textbf{0.9141} \\
InsectWingbeatSound & 0.5738 & 0.5911 & 0.4512 & 0.5715 & \textbf{0.6043} \\
MiddlePhalanxOutlineCorrect & 0.593 & 0.5233 & 0.6734 & \textbf{0.7808} & 0.6357 \\
NonInvasiveFatalECG\_Thorax1 & \textbf{0.9138} & 0.829 & 0.8263 & 0.8861 & 0.8328 \\
NonInvasiveFatalECG\_Thorax2 & \textbf{0.9227} & 0.855 & 0.845 & 0.9044 & 0.8673 \\
PhalangesOutlinesCorrect & \textbf{0.7986} & 0.6704 & 0.7345 & 0.7869 & 0.7443 \\
ProximalPhalanxOutlineAgeGroup & 0.8507 & 0.8166 & 0.7461 & 0.8341 & \textbf{0.8693} \\
ProximalPhalanxOutlineCorrect & \textbf{0.8454} & 0.754 & 0.7839 & 0.8419 & 0.7904 \\
ShapesAll & 0.715 & 0.5553 & 0.7084 & 0.808 & \textbf{0.8093} \\
StarLightCurves & \textbf{0.9635} & 0.9509 & 0.9573 & 0.963 & 0.9588 \\
Strawberry & \textbf{0.9602} & 0.6936 & 0.9161 & 0.9508 & 0.8957 \\
Two\_Patterns & 0.9552 & \textbf{0.9971} & 0.9944 & 0.9964 & 0.993 \\
UWaveGestureLibraryAll & \textbf{0.951} & 0.9277 & 0.7862 & 0.882 & 0.8778 \\
uWaveGestureLibrary\_X & \textbf{0.8202} & 0.7535 & 0.7136 & 0.7691 & 0.7636 \\
uWaveGestureLibrary\_Y & \textbf{0.7095} & 0.6601 & 0.5868 & 0.6557 & 0.649 \\
uWaveGestureLibrary\_Z & \textbf{0.7318} & 0.6639 & 0.6643 & 0.708 & 0.7006 \\
wafer & 0.9921 & 0.9882 & 0.9774 & \textbf{0.9966} & 0.995 \\
yoga & 0.7841 & 0.6757 & 0.6822 & \textbf{0.8077} & 0.7575 \\
\bottomrule
Avg & \textbf{0.8140} &  0.7531 & 0.7363 & 0.8126 & 0.7922 \\
\bottomrule
Wins & \textbf{13} & 1 & 0 & 6 & 4 \\
\bottomrule
\end{tabular}
\end{table*}

\begin{table*}[h]
\caption{Classification accuracies of semi-supervised methods with 20\% of labeled training set. \textbf{Bold} indicates best performance.}
\label{tab:ssl20}
\centering
\begin{tabular}{l|ccccc} 
\toprule
Dataset & TCGAN   & CA-TCC & SemiTime & SVM-TS2Vec   & LSVC-TS2Vec \\
\bottomrule
ChlorineConcentration & \textbf{0.6041} & 0.5068 & 0.5396 & 0.5483 & 0.5517 \\
Cricket\_X & 0.5297 & 0.4826 & 0.4565 & 0.5954 & \textbf{0.6077} \\
DistalPhalanxOutlineCorrect & \textbf{0.759} & 0.6773 & 0.7084 & 0.7464 & 0.7217 \\
ElectricDevices & 0.6187 & 0.6662 & 0.3016 & \textbf{0.6859} & 0.6632 \\
FordA & 0.891 & 0.8931 & 0.8937 & \textbf{0.925} & 0.9185 \\
FordB & \textbf{0.8829} & 0.8547 & 0.8769 & 0.7798 & 0.778 \\
HandOutlines & 0.841 & 0.8246 & 0.699 & 0.8941 & \textbf{0.907} \\
InsectWingbeatSound & 0.5241 & 0.5426 & 0.3514 & 0.5044 & \textbf{0.5525} \\
MiddlePhalanxOutlineCorrect & 0.6567 & 0.5033 & 0.6113 & \textbf{0.7808} & 0.5876 \\
NonInvasiveFatalECG\_Thorax1 & \textbf{0.8675} & 0.7264 & 0.7674 & 0.8295 & 0.7829 \\
NonInvasiveFatalECG\_Thorax2 & \textbf{0.8925} & 0.7563 & 0.7955 & 0.8663 & 0.8315 \\
PhalangesOutlinesCorrect & \textbf{0.7737} & 0.634 & 0.7357 & 0.7555 & 0.6991 \\
ProximalPhalanxOutlineAgeGroup & 0.8527 & 0.8078 & 0.7216 & 0.8351 & \textbf{0.8585} \\
ProximalPhalanxOutlineCorrect & \textbf{0.8378} & 0.7189 & 0.753 & 0.8144 & 0.7196 \\
ShapesAll & 0.6113 & 0.6097 & 0.5534 & 0.7203 & \textbf{0.728} \\
StarLightCurves & 0.9431 & 0.8853 & \textbf{0.965} & 0.9491 & 0.9429 \\
Strawberry & \textbf{0.9527} & 0.6943 & 0.8948 & 0.8784 & 0.8573 \\
Two\_Patterns & 0.8796 & 0.9546 & 0.9673 & \textbf{0.9902} & 0.9806 \\
UWaveGestureLibraryAll & \textbf{0.9322} & 0.8958 & 0.7358 & 0.8306 & 0.832 \\
uWaveGestureLibrary\_X & \textbf{0.7891} & 0.7059 & 0.6725 & 0.7178 & 0.7296 \\
uWaveGestureLibrary\_Y & \textbf{0.6773} & 0.6132 & 0.5455 & 0.6056 & 0.6013 \\
uWaveGestureLibrary\_Z & \textbf{0.7001} & 0.6267 & 0.6376 & 0.6788 & 0.6662 \\
wafer & \textbf{0.987} & 0.9726 & 0.9639 & 0.9865 & 0.986 \\
yoga & 0.712 & 0.6457 & 0.6355 & \textbf{0.7162} & 0.6899 \\
\bottomrule
Avg & \textbf{0.7798} & 0.7166 & 0.6993 & 0.7764 & 0.7581 \\
\bottomrule
Wins & \textbf{13} & 0 & 1 & 5 & 5 \\
\bottomrule
\end{tabular}
\end{table*}

\begin{table*}[h]
\caption{Classification accuracies of semi-supervised methods with 10\% of labeled training set. \textbf{Bold} indicates best performance.}
\label{tab:ssl10}
\centering
\begin{tabular}{l|ccccc} 
\toprule
Dataset & TCGAN   & CA-TCC & SemiTime & SVM-TS2Vec   & LSVC-TS2Vec \\
\bottomrule
ChlorineConcentration & \textbf{0.5524} & 0.484 & 0.4916 & 0.4882 & 0.5385 \\
Cricket\_X & 0.4374 & 0.3677 & 0.401 & 0.4482 & \textbf{0.4836} \\
DistalPhalanxOutlineCorrect & 0.6587 & 0.6433 & 0.664 & \textbf{0.7246} & 0.6906 \\
ElectricDevices & 0.5915 & 0.6405 & 0.2915 & \textbf{0.6667} & 0.647 \\
FordA & 0.8695 & 0.8764 & 0.8819 & \textbf{0.9197} & 0.905 \\
FordB & \textbf{0.8763} & 0.8149 & 0.832 & 0.782 & 0.7728 \\
HandOutlines & 0.8186 & 0.8176 & 0.6068 & 0.8935 & \textbf{0.9043} \\
InsectWingbeatSound & \textbf{0.483} & 0.4809 & 0.2772 & 0.4444 & 0.4738 \\
MiddlePhalanxOutlineCorrect & 0.6147 & 0.4713 & 0.5918 & \textbf{0.6948} & 0.5581 \\
NonInvasiveFatalECG\_Thorax1 & \textbf{0.8063} & 0.5976 & 0.6807 & 0.7421 & 0.7063 \\
NonInvasiveFatalECG\_Thorax2 & \textbf{0.8417} & 0.6559 & 0.7113 & 0.8004 & 0.7702 \\
PhalangesOutlinesCorrect & \textbf{0.7364} & 0.6179 & 0.6832 & 0.7354 & 0.6685 \\
ProximalPhalanxOutlineAgeGroup & 0.8371 & 0.8107 & 0.7312 & 0.7951 & \textbf{0.8478} \\
ProximalPhalanxOutlineCorrect & \textbf{0.8027} & 0.7285 & 0.6918 & 0.7677 & 0.7065 \\
ShapesAll & 0.4957 & 0.497 & 0.4037 & 0.6177 & \textbf{0.6377} \\
StarLightCurves & 0.9373 & 0.9004 & \textbf{0.973} & 0.9346 & 0.9236 \\
Strawberry & \textbf{0.9044} & 0.6998 & 0.8719 & 0.8578 & 0.82 \\
Two\_Patterns & 0.7892 & 0.9002 & 0.9001 & \textbf{0.9654} & 0.954 \\
UWaveGestureLibraryAll & \textbf{0.891} & 0.8784 & 0.6943 & 0.7513 & 0.7639 \\
uWaveGestureLibrary\_X & \textbf{0.751} & 0.6726 & 0.6126 & 0.6759 & 0.6838 \\
uWaveGestureLibrary\_Y & \textbf{0.627} & 0.5973 & 0.4734 & 0.5598 & 0.5512 \\
uWaveGestureLibrary\_Z & \textbf{0.6473} & 0.5976 & 0.5887 & 0.6442 & 0.64 \\
wafer & \textbf{0.9731} & 0.9687 & 0.9592 & 0.9722 & 0.9687 \\
yoga & \textbf{0.6357} & 0.5937 & 0.5797 & 0.6279 & 0.6058 \\
\bottomrule
Avg & \textbf{0.7324} & 0.6797 & 0.6497 & 0.7296 & 0.7176 \\
\bottomrule
Wins & \textbf{14} & 0 & 1 & 5 & 4 \\
\bottomrule
\end{tabular}
\end{table*}

First, we compare LR-TCGAN with advanced single-supervised CNNs, i.e., the FCN and ResNet in \cite{ismail2019deep}. We consider the FordA and wafer datasets because both datasets are large enough and all classifiers obtain competitive accuracy on the complete training set. We run each setting 5 times with different random seeds and report the average accuracy on the default test split.
Fig.~\ref{fig:semi-supervised} presents the results. A subplot corresponds to one dataset, x-axis denotes the proportion of labeled training data used for supervised learning, and y-axis is the average accuracy. 
As expected, with a decreasing proportion of available labels, all classifiers consistently get worse, but the decay of fully supervised networks (ResNet and FCN) is significantly greater than that of TCGAN classifiers with unsupervised pretraining. Moreover, ResNet and FCN tend to be unstable on the imbalanced dataset wafer. 

Moreover, we compare LR-TCGAN with two leading semi-supervised models: \textbf{CA-TCC} \cite{eldele2022self} and Semi-supervised Time series classification architecture (\textbf{SemiTime}) \cite{fan2021semi}.
CA-TCC has been introduced in Section \ref{sec:tcgan-tsc}. SemiTime ingests labeled and unlabeled data simultaneously. Specifically, SemiTime conducts supervised classification directly on the labeled data. To utilize the unlabeled data, SemiTime proposes a self-supervised predictor that samples segments of past-future pair from time series and predicts the temporal relation between them. We reproduce SemiTime using the open source code \footnote{{SemiTime: \url{https://github.com/haoyfan/SemiTime}}}. 
Additionally, we evaluate the top two classifiers associated with TS2Vec \cite{yue2021ts2vec}, i.e., \textbf{SVM-TS2Vec} and \textbf{LSVC-TS2Vec}, in the semi-supervised scenario. Please refer to Section \ref{sec:tcgan-tsc} for more details on TS2Vec.

Our experiments encompass 24 UCR datasets. Specifically, we rank the 85 UCR datasets by the average number of samples in each class and select the top 15 datasets. Besides, we include 9 datasets from competitors \cite{fan2021semi,eldele2022self}. 

We train semi-supervised models with different proportions of labeled training data and report the accuracy on the test data.
Tab.~\ref{tab:ssl40}, Tab.~\ref{tab:ssl20} and Tab.~\ref{tab:ssl10} present the results, with each entry representing the average accuracy over 5 random runs. TCGAN is generally the best. Specifically, TCGAN wins 13, 13, and 14 out of 24 datasets with respect to 40\%, 20\%, and 10\% of labeled training data, respectively. 

In summary, the above results demonstrate that TCGAN classifiers can make use of unlabeled data to obtain a decent and stable accuracy even when the labeled data are highly limited.

\subsubsection{TCGAN Representations for Imbalanced Classification}
\label{sec:tcgan-tsc-extreme-imb}

We explore the effectiveness of LR-TCGAN and advanced single-supervised CNNs (i.e., the FCN and ResNet in \cite{ismail2019deep}) for imbalanced classification. We include 5 UCR datasets with imbalance ratios greater than 30. ImBalance Ratio (IBRa) is the ratio between the number of samples from the most and least frequent classes. We use the common imbalanced classification evaluation metric, the weighted F1 score.  
The results in Tab.\ref{tab:imb} show that LR-TCGAN achieves the best results on all datasets.

\begin{table*}[h]
\caption{Mean and standard deviation (in bracket) of weighted F1 scores on imbalanced datasets. \textbf{Bold} indicates best performance.}
\label{tab:imb}
\centering
\begin{tabular}{l|c|ccc} 
\toprule
Dataset & IBRa & FCN  & ResNet  & LR-TCGAN \\
\midrule
50words &  52.000 &  0.5850(0.0146) &  0.7054(0.0228) &  \textbf{0.7592(0.0094)} \\
ECG5000 &  146.000 &  0.9300(0.0065) &  0.9281(0.0037) &  \textbf{0.9324(0.0010)} \\
MedicalImages & 33.833 & 0.7211(0.0718) & 0.7625(0.0143) & \textbf{0.7699(0.0144)} \\
ProximalPhalanxTW &  36.000 &  0.7565(0.0439) &  0.7682(0.0118) &  \textbf{0.7825(0.0124)} \\
WordsSynonyms & 30.000 & 0.4683(0.0510) & 0.5872(0.0173) & \textbf{0.6650(0.0117)} \\
\bottomrule
\end{tabular}
\end{table*}

\subsection{TCGAN Representations Metric Space for Clustering}
\label{sec:tcgan-repr-space}

Effective clustering requires a representation space that defines a meaningful measure between different samples and uncovers discriminative structures inherent in the data. 
We will demonstrate that TCGAN transformation can enable a simple clustering method to achieve superior performance. 

We make comparisons with the following models: 
\begin{itemize}
  \item \textbf{k-means} with Euclidean distance and raw data input is a common clustering method.   
  \item \textbf{k-means-TCGAN}. Our model feeds TCGAN representations to a k-means with Euclidean distance.
  \item \textbf{k-shape} \cite{paparrizos2017fast} is a superior raw-data-based algorithm that uses a normalized version of the cross-correlation as a distance measure to cluster raw data.
  \item \textbf{k-means-SPIRAL-DTW} \cite{lei2019similarity} is an advanced feature-based method. It trains a k-means algorithm with DTW distance on the extracted SPIRAL features. 
\end{itemize}
We implement k-means with scikit-learn, run k-shape with the public source code \footnote{{k-shape source code: \url{https://tslearn.readthedocs.io/en/stable/gen_modules/clustering/tslearn.clustering.KShape.html}}}, and directly copy the result of k-means-SPIRAL-DTW from the primitive paper due to unavailability of the source code. All experiments are run 5 times with different random seeds. The k-means with raw data input is deterministic and thus is run once. 

Tab.~\ref{tab:clus-avg} reports the average of clustering NMIs on all UCR 85 datasets. Our method, k-means-TCGAN, achieves the best results, even though k-means-SPIRAL-DTW applies a more complicated distance measure, DTW. 
Furthermore, we observe the pairwise comparison results of k-means-TCGAN, k-means and k-shape. k-means-SPIRAL-DTW is excluded because the detailed results for each dataset are not available. In Tab.~\ref{tab:clus-pair},  columns ">", "=", "<" denote the number of datasets over which the active competitor is better, equal, or worse, respectively, in comparison with the other. Column "$p$(WSRT)" presents the $p$-values of the Wilcoxon signed rank test (WSRT).   
The results show that k-means-TCGAN is significantly better than k-means and k-shape ($p$(WSRT) $<0.05$). 
It should be noted that the k-means algorithm in scikit-learn uses k-means++ for initialization, while k-shape uses an inferior initialization method (random initialization). Therefore, in our experiments, k-shape does not significantly outperform k-means. In fact, it is not trivial to use k-means++ in k-shape. From this perspective, an independent encoder is more flexible for use in conjunction with many advanced clustering algorithms.

We also apply t-SNE \cite{maaten2008visualizing} to find a two-dimensional embedding and inspect the representation space. Fig.~\ref{intro-case} visualizes the Two\_Patterns dataset where instances of different classes become more distinguishable in the representation space. As a side effect, such a t-SNE plot could be an exploratory data analysis technique to help domain experts understand time series as in \cite{yeh2016matrix}. 

\begin{table*}[h]
\caption{Average of clustering NMIs on 85 UCR datasets. \textbf{Bold} indicates best performance.}
\label{tab:clus-avg}
\centering
\begin{tabular}{l|cccc}  
\hline
        & k-means-TCGAN    & k-means   & k-shape    & k-means-SPIRAL-DTW \\
\hline
NMI     & \textbf{0.3412}   & 0.2889   & 0.3097   & 0.332\\
\bottomrule
\end{tabular}
\end{table*}

\begin{table}[h]
\caption{Pairwise comparison of clustering methods on 85 UCR datasets.}
\label{tab:clus-pair}
\centering
\begin{tabular}{l|cccc}  
\hline
        & >    &=   & <    & $p$(WSRT) \\
\hline
k-means-TCGAN vs. k-means  & 64   &2   & 19   & 5.64E-07\\
\hline
k-means-TCGAN vs. k-shape     & 51   &0   & 34    & 0.0149 \\
\hline
k-shape vs. k-means          & 48   &0   & 37    & 0.1551 \\
\bottomrule
\end{tabular}
\end{table}

\subsection{Runtime}
\label{sec:runtime}

We report the total runtime over 85 UCR datasets. 
Training TCGAN takes 8.6834 hours. 
Extracting features requires 38.9938 seconds.
SM-TCGAN, LR-TCGAN and LSVC-TCGAN classifiers spend 2.9955 hours, 0.1111 hours, and 0.0913 hours, respectively.
Clustering with kmeans takes 0.1332 hours.
In addition, we have discussed the runtime of time-series GANs in Section~\ref{sec:gans-comp}, particularly Tab.~\ref{tab:gen}.
Similar to existing DNNs, training neural networks (i.e., TCGAN and SM classifier) with enough epochs takes the most time. It is reassuring that TCGAN only needs to be trained once for all downstream applications.

\section{Conclusion}
\label{sec:con}
We introduced TCGAN for time series classification and clustering. TCGAN is trained by playing an adversarial game in absence any labeled information, and then parts of the trained TCGAN are reused to construct a representation encoder for linear classification and clustering methods.
Our extensive experiments on synthetic and real-world time series datasets demonstrate that TCGAN outperforms existing time-series GANs in terms of effectiveness and efficiency. 
TCGAN representations enable simple classifiers to achieve higher accuracy and stability than both leading unsupervised and supervised CNNs even in highly limited and/or imbalanced labeled data scenarios. 
Plus, the pairwise similarities between time series are well preserved in TCGAN transformation, making the distance-based clustering method more effective. 
We hope our work will motivate further research on applying GANs or generative models to address the shortage of labeled time series data. 

\printcredits

\bibliographystyle{cas-model2-names}

\bibliography{cas-refs}



\end{document}